\documentclass[acmsmall,authorversion]{acmart}

\acmJournal{PACMPL}
\acmVolume{1}
\acmNumber{CONF} 
\acmArticle{1}
\acmYear{2018}
\acmMonth{1}
\acmDOI{} 
\startPage{1}

\setcopyright{none}

\usepackage[english]{babel}
\usepackage[T1]{fontenc}
\usepackage{changes}
\usepackage{times}
\usepackage{mathtools}
\usepackage{amsmath}
\usepackage{amsthm}
\usepackage[dvipsnames]{xcolor}
\usepackage{enumerate}
\usepackage{graphicx}
\usepackage[export]{adjustbox}
\usepackage{booktabs}
\usepackage{tikz, pgfplots}
\usepackage{listings}
\usepackage{subcaption}
\usepackage[normalem]{ulem}
\usepackage{doi}
\useunder{\uline}{\ul}{}
\usepackage[toc,page]{appendix}
\usepackage{paralist}
\usepackage{cleveref}
\usepackage{scalerel}
\usepackage{tabu}
\usepackage[width=5em, filledcolor=black, emptycolor=white]{progressbar}

\usepackage[textfont=small,font=small,labelfont=bf]{caption}

\definechangesauthor[name={EY}, color=blue]{yahave}
\setremarkmarkup{(#2)}

\newcommand*{\origrightarrow}{}

\renewcommand*{\textrightarrow}{\fontfamily{cmr}\selectfont\origrightarrow}

\usepackage{ifthen}

\newcommand{\todolow}[1]{}

\newcommand{\COMMENT}[1]{}
\newcommand{\ignore}[1]{}

\newcommand{\figref}[1]{Fig.~\ref{Fi:#1}}

\newcommand{\secref}[1]{Section~\ref{Se:#1}}

\newcommand{\figlabel}[1]{\label{Fi:#1}}

\newcommand{\seclabel}[1]{\label{Se:#1}}

\theoremstyle{definition}
\newtheorem{definition}{Definition}[]

\newboolean{TR}
\setboolean{TR}{false}
\ifthenelse{\boolean{TR}}{

\newcommand{\TrOnly}[1]{#1}
\newcommand{\SubOnly}[1]{}
\newcommand{\TrOnlyInFootnote}[1]{#1}
\newcommand{\TrOnlyInTable}[1]{#1}}
{

\newcommand{\TrOnly}[1]{}
\newcommand{\SubOnly}[1]{#1}
\newcommand{\TrOnlyInFootnote}[1]{}
\newcommand{\TrOnlyInTable}[1]{}}

\newcommand{\scode}[1]{{\small \texttt{#1}}}

\definecolor{airforceblue}{rgb}{0.36, 0.54, 0.66}
\definecolor{awesome}{rgb}{1.0, 0.13, 0.32}
\definecolor{blue(pigment)}{rgb}{0.2, 0.2, 0.6}
\definecolor{britishracinggreen}{rgb}{0.0, 0.26, 0.15}
\definecolor{cadmiumgreen}{rgb}{0.0, 0.42, 0.24}
\definecolor{carminered}{rgb}{1.0, 0.0, 0.22}
\definecolor{electricindigo}{rgb}{0.44, 0.0, 1.0}
\definecolor{mygray}{rgb}{0.5,0.5,0.5}
\hypersetup{colorlinks=true,linkcolor=black, citecolor=black, urlcolor=black}

\definecolor{javared}{rgb}{0.6,0,0} 
\definecolor{javagreen}{rgb}{0.25,0.5,0.35} 
\definecolor{javapurple}{rgb}{0.5,0,0.35} 
\definecolor{javadocblue}{rgb}{0.25,0.35,0.75} 

\definecolor{path-orange}{HTML}{D79B00}
\definecolor{path-green}{HTML}{517040}

\usepackage{listings}
\newcommand{\rulesep}{\unskip\ \vrule\ }

\begin{document}

\title{code2vec: Learning Distributed Representations of Code}        


\author{Uri Alon}
\affiliation{
  \institution{Technion}            
}
\email{urialon@cs.technion.ac.il}          

\author{Meital Zilberstein}
\affiliation{
  \institution{Technion}           
}
\email{mbs@cs.technion.ac.il}         

\author{Omer Levy}
\affiliation{
  \institution{Facebook AI Research}           
}
\email{omerlevy@gmail.com}         

\author{Eran Yahav}
\affiliation{
  \institution{Technion}           
}
\email{yahave@cs.technion.ac.il}         


\begin{abstract}

We present a neural model for representing snippets of code as continuous distributed vectors (``code embeddings''). The main idea is to represent a code snippet as a single fixed-length \emph{code vector}, which can be used to predict semantic properties of the snippet. This is performed by decomposing code to a collection of paths in its abstract syntax tree, and learning the atomic representation of each path \emph{simultaneously} with learning how to aggregate a set of them.

We demonstrate the effectiveness of our approach by using it to predict a method's name from the vector representation of its body. We evaluate our approach by training a model on a dataset of $14$M methods. We show that code vectors trained on this dataset can predict method names from files that were completely unobserved during training. Furthermore, we show that our model learns useful method name vectors that capture semantic similarities, combinations, and analogies.

Comparing previous techniques over the same data set, our approach obtains a relative improvement of over $75\%$, being the first to successfully predict method names based on a large, cross-project, corpus. Our trained model, visualizations and vector similarities are available as an interactive online demo at \url{http://code2vec.org}. The code, data and trained models are available at \url{https://github.com/tech-srl/code2vec}.
\end{abstract}



\maketitle

\section{Introduction}\seclabel{Intro}

Distributed representations of words (such as ``word2vec'') \cite{mikolovEfficient2013, mikolovDistributed2013, glove2014}, sentences, paragraphs, and documents (such as ``doc2vec'') \cite{doc2vec} played a key role in unlocking the potential of neural networks for natural language processing (NLP) tasks \cite{bengio2003, collobert2008, SocherEtAl2011:RNN, turian2010, glorot2011domain, turney2006similarity}. Methods for learning distributed representations produce low-dimensional vector representations for objects, referred to as \emph{embeddings}. In these vectors, the ``meaning'' of an element is distributed across multiple vector component , such that semantically similar objects are mapped to close vectors.

\paragraph{Goal:} The goal of this paper is to learn \emph{code embeddings}, continuous vectors for representing snippets of code. By learning code embeddings, our long-term goal is to enable the application of neural techniques to a wide-range of programming-languages tasks. In this paper, we use the motivating task of \emph{semantic labeling of code snippets}.


\begin{figure}[t]
\hspace{-4mm}
\begin{tabular}{lr}
\begin{subfigure}[h]{0.6\textwidth}
\vskip 0pt
\fbox{\includegraphics[scale=0.25]{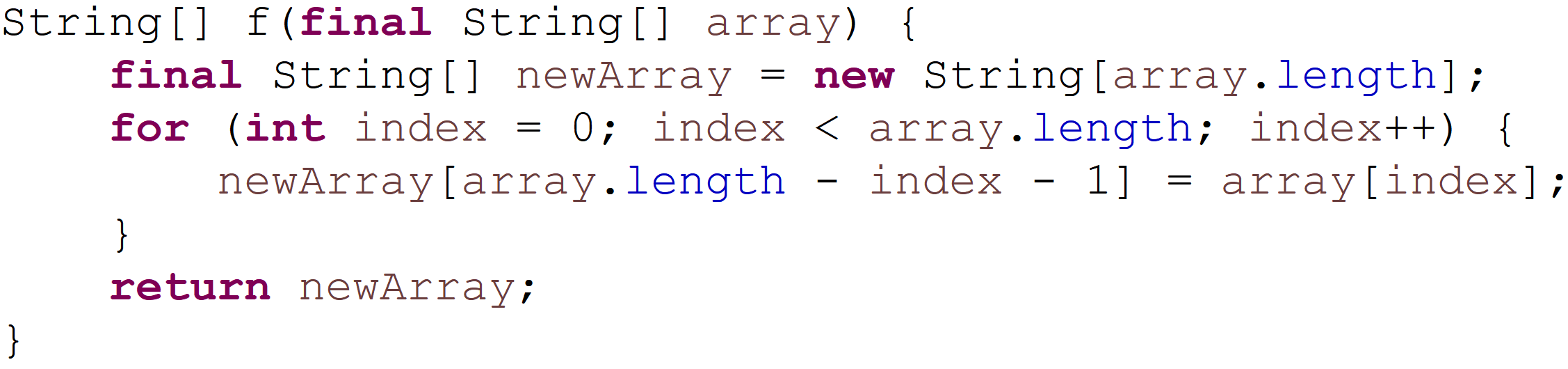}}
\end{subfigure}
\hspace{5mm}
\begin{subfigure}[h]{0.35\textwidth}
\small
\begin{tabular}{lll}
{\ul Predictions}  & &  \\
\textbf{reverseArray}   &\progressbar{0.7734}& $77.34\%$      \\
\textbf{reverse}        &\progressbar{0.1818}& $18.18\%$ \\
\textbf{subArray}       &\progressbar{0.0145}& $1.45\%$ \\
\textbf{copyArray}      &\progressbar{0.0074}& $0.74\%$ \\
\end{tabular}
\end{subfigure}

\end{tabular}
\caption{A code snippet and its predicted labels as computed by our model.}
\label{reverse-nopaths}
\end{figure}

\paragraph{Motivating task: semantic labeling of code snippets}
Consider the method in \Cref{reverse-nopaths}. The method contains only low-level assignments to arrays, but a human reading the code may (correctly) label it as performing the \emph{reverse} operation. Our goal is to predict such labels automatically.
The right hand side of \Cref{reverse-nopaths} shows the labels predicted automatically using our approach. The most likely prediction ($77.34\%$) is \emph{reverseArray}. \Cref{Evaluation} provides additional examples.

Intuitively, this problem is hard because it requires \emph{learning a correspondence} between the \emph{entire content of a method} and a semantic label. That is, it requires aggregating possibly hundreds of expressions and statements from the method body into a single, descriptive label.

\paragraph{Our approach} we present a novel framework for predicting program properties using neural networks. Our main contribution is a neural network that learns \emph{code embeddings} - continuous distributed vector representations for code. The code embeddings allow us to model correspondence between code snippet and labels in a natural and effective manner.

Our neural network architecture uses a representation of code snippets that leverages the structured nature of source code, and learns to aggregate multiple syntactic paths into a single vector. This ability is fundamental for the application of deep learning in programming languages. By analogy, word embeddings in natural language processing (NLP) started a revolution of application of deep learning for NLP tasks.

The input to our model is a code snippet and a corresponding tag, label, caption, or name. This label expresses the semantic property that we wish the network to model, for example: a tag that should be assigned to the snippet, or the name of the method, class, or project that the snippet was taken from.
Let $\mathcal{C}$ be the code snippet and $\mathcal{L}$ be the corresponding label or tag. Our underlying hypothesis is that \emph{the distribution of labels can be inferred from syntactic paths in $\mathcal{C}$}. Our model therefore attempts to learn the label distribution, conditioned on the code:
$P\left(\mathcal{L}|\mathcal{C}\right)$.

We demonstrate the effectiveness of our approach for the task of predicting a method's name given its body.
This problem is important as good method names make code easier to understand and maintain. A good name for a method provides a high-level summary of its purpose. Ideally, \emph{``If you have a good method name, you don't need to look at the body.''}~\cite{fowler1999refactoring}. Choosing good names can be especially critical for methods that are part of public APIs, as poor method names can doom a project to irrelevance \cite{allamanis2015, Host:2009:DMN:1615184.1615204}.

\ignore{
\paragraph{Predicting method names}
We demonstrate our approach on the task of predicting method names in a program. Given a method body, we use the network to predict a descriptive name. Predicting names for program elements can improve code quality by making sure that elements are given descriptive names that capture their essence~\cite{Lawrie:2006:WNS:1135772.1136161, takang96, liblit2006, allamanis2015, Host:2009:DMN:1615184.1615204}.
}


\begin{sloppypar}
\paragraph{Capturing semantic similarity between names}
During the process of learning code vectors, a parallel vocabulary of vectors of the labels is learned. When using our model to predict method names, the method-name vectors provide surprising semantic similarities and analogies. For example,  $vector(\scode{equals}) + vector(\scode{toLowerCase})$ results in a vector that is closest to $vector(\scode{equalsIgnoreCase})$.
\end{sloppypar}

Similar to the famous NLP example of: $vec(``king'') - vec(``man'') + vec(``woman'') \approx vec(``queen'')$ \cite{mikolov2013linguistic}, our model learns analogies that are relevant to source code, such as: \emph{``\scode{receive} is to \scode{send} as \scode{download} is to: \underline{\scode{upload}}''}. \Cref{similarities} shows additional examples, and \Cref{qualitative} provides a detailed discussion.

\begin{table}[]
\footnotesize
\begin{tabu}{|l|l|[1pt]|[1pt]l|l|}
\hline
A  & $\approx$B  & A  & $\approx$B            \\ \hline
size     & getSize,     length,     getCount,   getLength   & executeQuery     & executeSql, runQuery, getResultSet \\
active   & isActive, setActive, getIsActive, enabled                 & actionPerformed & itemStateChanged, mouseClicked, keyPressed\\
done     & end, stop, terminate & toString & getName, getDescription, getDisplayName \\
toJson   & serialize, toJsonString, getJson, asJson, & equal   & eq, notEqual, greaterOrEqual, lessOrEqual \\
   run      & execute, call, init, start & error    & fatalError, warning, warn\\
\hline
\end{tabu}
\centering
\caption{Semantic similarities between method names.}
\label{similarities}
\end{table}

\subsection{Applications}
Embedding a code snippet as a vector has a variety of machine-learning based applications, since machine-learning algorithms usually take vectors as their inputs. Such direct applications, that we examine in this paper, are:
\begin{enumerate}
\item \emph{Automatic code review} -  suggesting better method names when the name given by the developer doesn't match the method's functionality, thus preventing naming bugs, improving the readability and maintenance of code, and easing the use of public APIs. This application was previously shown to be of significant importance \cite{fowler1999refactoring, allamanis2015, Host:2009:DMN:1615184.1615204}.
\item \emph{Retrieval and API discovery} - semantic similarities enable search in ``the problem domain'' instead of search ``in the solution domain''. For example, a developer might look for a \scode{serialize} method, while the equivalent method of the class is named \scode{toJson} as serialization is performed via \scode{json}. An automatic tool that looks for the most similar \emph{vector} to the requested name among the available methods will find \scode{toJson} (\Cref{similarities}). Such semantic similarities are difficult to find without our approach. Further, an automatic tool which uses our vectors can easily notice that a programmer is using the method \scode{equals} right after \scode{toLowerCase} and suggest to use \scode{equalsIgnoreCase} instead (\Cref{additions}).
\end{enumerate}

The code vectors we produce can be used as input to any machine learning pipeline that performs tasks such as code retrieval, captioning, classification and tagging, or as a metric for measuring similarity between snippets of code for ranking and clone detection. The novelty of our approach is in its ability to produce vectors that capture properties of snippets of code, such that similar snippets (according to any desired criteria) are assigned with similar vectors. This ability unlocks a variety of applications for working with machine-learning algorithms on code, since machine learning algorithms usually take vectors as their input, just as word2vec unlocked a wide range of applications and improved almost every NLP application.

We deliberately picked the difficult task of method name prediction, for which prior results were low \cite{conv16, pigeon-pldi, allamanis2015} as an evaluation benchmark.
Succeeding in this challenging task implies good performance in other tasks such as: predicting whether or not a program performs I/O, predicting the required dependencies of a program, and predicting whether a program is a suspected malware.
We show that even for this challenging benchmark, our technique provides dramatic improvement over previous works.

\subsection{Challenges: Representation and Attention}
Assigning a semantic label to a code snippet (such as a name to a method) is an example for a class of problems that require a compact semantic descriptor of a  snippet. The question is how to represent code snippets in a way that captures some semantic information, is reusable across programs, and can be used to predict properties such as a label for the snippet. This leads to two  challenges:
\begin{compactitem}
\item Representing a snippet in a way that enables learning across programs.
\item Learning which parts in the representation are relevant to prediction of the desired property, and learning what is the order of importance of each part.
\end{compactitem}

\paragraph{Representation}
NLP methods typically treat text as a linear sequence of tokens. Indeed, many existing approaches also represent source code as a token stream \cite{allamanis2014-conventions, allamanis2013, conv16, movshovitz2013natural, white2015, Hindle:2012:NS:2337223.2337322}. However, as observed previously~\cite{phog16, jsnice2015, pigeon-pldi}, programming languages can greatly benefit from representations that leverage the structured nature of their syntax.

We note that there is a tradeoff between the degree of program-analysis required to extract the representation, and the learning effort that follows. Performing no program-analysis at all, and learning from the program's surface text, incurs a significant learning effort that is often prohibitive in the amounts of data required. Intuitively, this is because the learning model has to re-learn the syntax and semantics of the programming language from the data. On the other end of the spectrum, performing a deep program-analysis to extract the representation may make the learned model language-specific (and even task-specific).

Following previous works~\cite{pigeon-pldi, jsnice2015}, we use paths in the program's abstract syntax tree (AST) as our representation. By representing a code snippet using its syntactic paths, we can capture regularities that reflect common code patterns. We find that this representation significantly lowers the learning effort (compared to learning over program text), and is still scalable and general such that it can be applied to a wide range of problems and large amounts of code.

We represent a given code snippet as a bag (multiset) of its extracted paths. The challenge is then \emph{how to aggregate a bag of contexts, and which paths to focus on for making a prediction}.

\paragraph{Attention}
Intuitively, the problem is to learn a correspondence between a bag of path-contexts and a label. Representing each bag of path-contexts \emph{monolithically} is going to suffer from sparsity -- even similar methods will not have the \emph{exact} same bag of path-contexts. We therefore need a \emph{compositional} mechanism that can aggregate a bag of path-contexts such that bags that yield the same label are mapped to close vectors. Such a compositional mechanism would be able to generalize and represent new unseen bags by leveraging the fact that during training it observed the individual path-contexts and their components (paths, values, etc.) as parts of other bags.

To address this challenge, which is the focus of this paper, we use a novel attention network architecture. Attention models have gained much popularity recently, mainly for neural machine translation (NMT) \cite{bahdanau14, luong15, vaswani2017attention}, reading comprehension \cite{levy2017zeroshot, seo2016bidirectional}, image captioning \cite{xu2015}, and more \cite{chorowski2015attention, bahdanau2016end, mnih2014, ba2014}.

In our model, neural attention learns how much focus (``attention'') should be given to each element in a bag of path-contexts. It allows us to precisely aggregate the information captured in each individual path-context into a single vector that captures information about the entire code snippet. As we show in \Cref{qualitative}, our model is relatively interpretable: the weights allocated by our attention mechanism can be visualized to understand the relative importance of each path-context in a prediction. The attention mechanism is \emph{learned simultaneously with the embeddings, optimizing both the atomic representations of paths and the ability to compose multiple contexts into a single code vector}.



\ignore{
While these models are sometimes very costly in terms of training time, our neural architecture is simple enough to scale and train on millions of examples.
}

\ignore{
    Each of the three components in a path-context $(x_{s},p,x_t)$ is assigned a real vector representation, or an \emph{embedding}. The three vectors of the tuple are concatenated to represent a path-context embedding. Given a set of path-context embeddings (representing the bag of path contexts), an attention mechanism learns weights on how these embeddings should be taken into a weighted average.
}

\paragraph{Soft and hard attention} The terms ``soft'' and ``hard'' attention were proposed for the task of image caption generation by \citet{xu2015}. Applied in our setting, \emph{soft-attention} means that weights are distributed ``softly'' over all path-contexts in a code snippet, while \emph{hard-attention} refers to selection of a single path-context to focus on at a time.
The use of \emph{soft-attention} over syntactic paths is the main understanding that provides this work much better results than previous works. We compare our model with an equivalent model that uses hard-attention in \Cref{alternative}, and show that \emph{soft-attention} is more efficient for modeling code.

\subsection{Existing Techniques}

The problem of predicting program properties by learning from big code has seen tremendous interest and progress in recent years~\cite{decisionTrees2016,allamanis2014-conventions,phog16,allamanis2013,Hindle:2012:NS:2337223.2337322}. The ability to predict semantic properties of a program without running it, and with little or no semantic analysis at all, has a wide range of applications: predicting names for program entities~\cite{pigeon-pldi,jsnice2015,allamanis2015}, code completion~\cite{raychev14,Mishne12}, code summarization~\cite{conv16}, code generation \cite{murali2018bayou,maddison2014,amodio2017neural,lu2017data}, and more (see~\cite{allamanis2017survey,VY16} for a survey).

A recent work \cite{pigeon-pldi} used syntactic paths with Conditional Random Fields (CRFs) for the task of predicting method names in Java. Our work achieves significantly better results for the same task on the same dataset: F1 score of $58.4$ vs. $49.9$ (a relative improvement of $17\%$), while training $5X$ faster thanks to our ability to use a GPU, which cannot be used in their model. Further, their approach can only perform predictions for the exact task that it was trained for, while our approach produces \emph{code vectors} that once trained for a single task, are useful for other tasks as well.
In \Cref{Parameters} we discuss more conceptual advantages compared to the model of \citet{pigeon-pldi} which are generalization ability and a reduction of \emph{polynomial} space complexity with \emph{linear} space.

Distributed representations of code identifiers were first suggested by \citet{allamanis2015}, and used to predict variable, method, and class names based on token context features.
\citet{conv16} were also the first to consider the problem of predicting method names. Their technique used a Convolutional Neural Network (CNN) where locality in the model is based on textual locality in source code. While their technique works well when training and prediction are performed within the scope of the same project, they report poor results when used across different projects (as we reproduce in \Cref{quantitative}). Thus, the problem of predicting method names based on a large corpus has remained an open problem until now. To the best of our knowledge, our technique is the first to train an effective cross-project model for predicting method names.


\subsection{Contributions}

The main contributions of this paper are:
\begin{itemize}
\item A path-based attention model for learning vectors for arbitrary-sized snippet of code. This model allows to embed a program, which is a discrete object, into a continuous space, such that it can be fed into a deep learning pipeline for various tasks.
\item As a benchmark for our approach, we perform a quantitative evaluation for predicting cross-project method names, trained on more than $14$M methods of real-world data, and compared with previous works. Experiments show that our approach achieves significantly better results than previous works which used Long Short-Term Memory networks (LSTMs), CNNs and CRFs.
\item A qualitative evaluation that interprets the attention that the model has learned to give to the different path-contexts when making predictions.
\item A collection of method name embeddings, which often assign semantically similar names to similar vectors, and even allows to compute analogies using simple vector arithmetic.
\item An analysis that shows the significant advantages in terms of generalization ability and space complexity of our model, compared to previous non-neural works such as \citet{pigeon-pldi} and \citet{jsnice2015}.
\end{itemize}

\section{Overview}\seclabel{Overview}

\begin{figure*}[]
\centering
\begin{minipage}{7in}
\hspace{-0.4in}
\begin{tabular}{lll}
\\
\begin{subfigure}[t]{0.28\textwidth}
\includegraphics[scale=0.07]{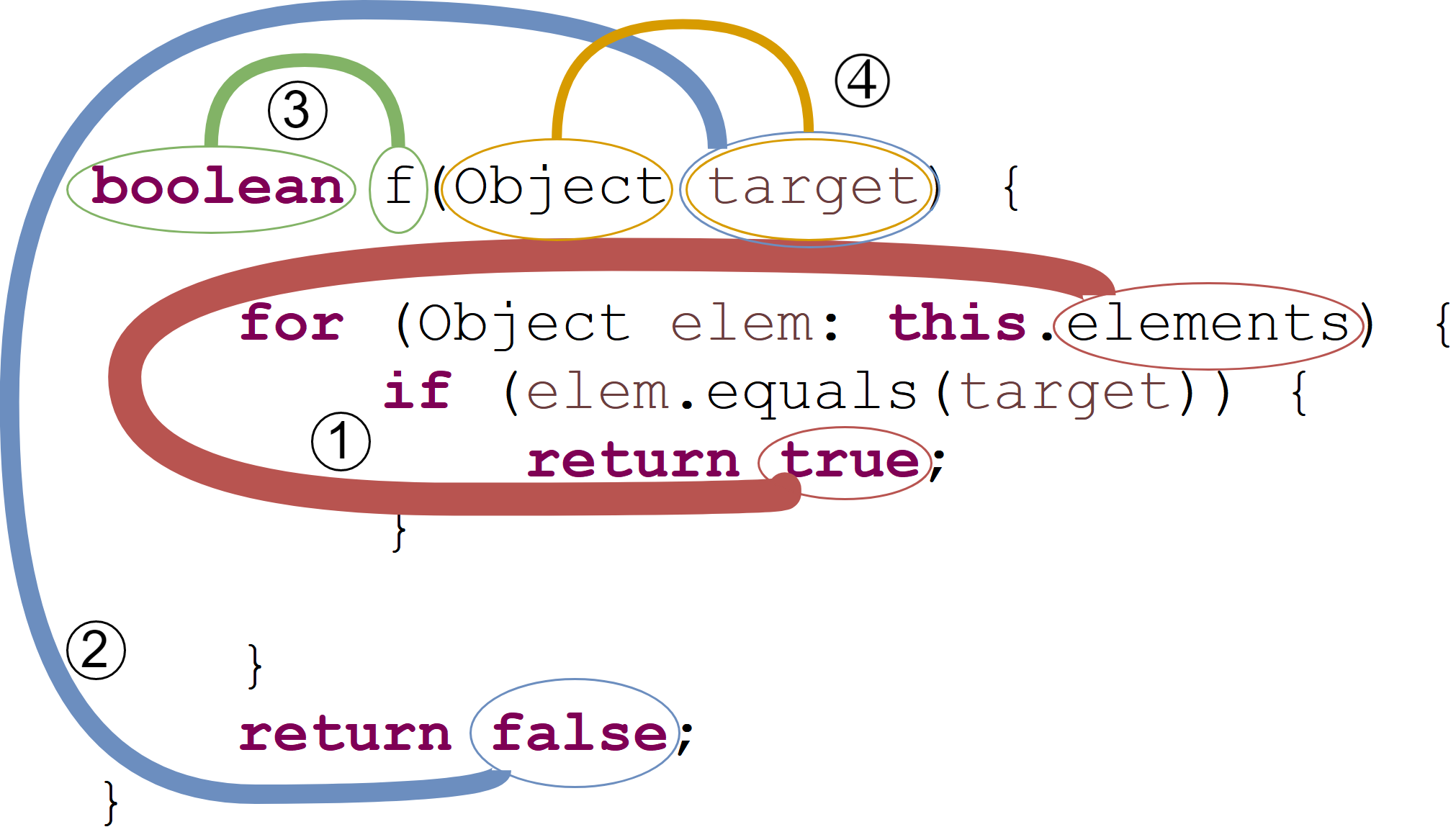}
\caption{}
\label{overview-a}
\vspace{1mm}
\end{subfigure}
\rulesep
\begin{subfigure}[t]{0.32\textwidth}
\includegraphics[scale=0.07]{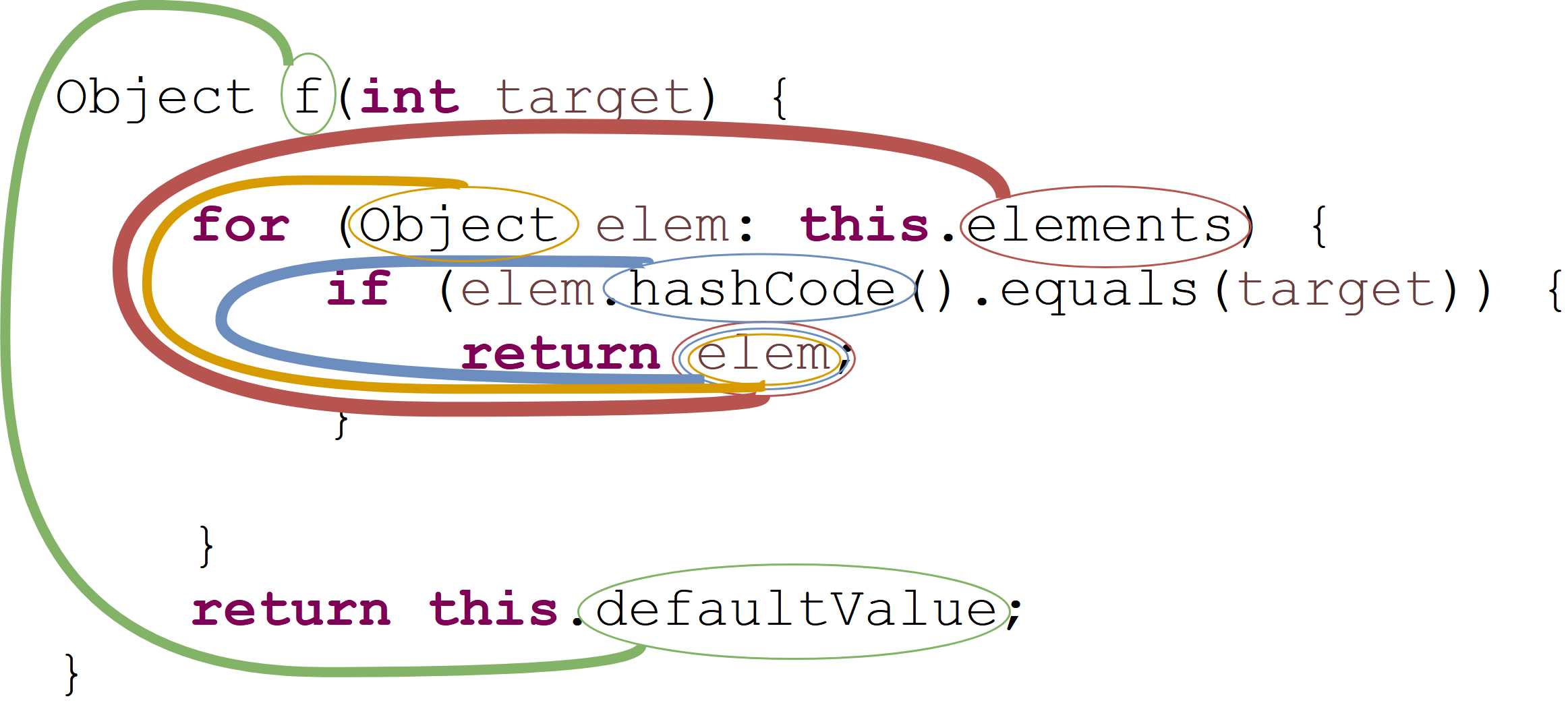}
\caption{}
\label{overview-b}
\vspace{1mm}
\end{subfigure}
\rulesep
\begin{subfigure}[t]{0.3\textwidth}
\includegraphics[scale=0.07]{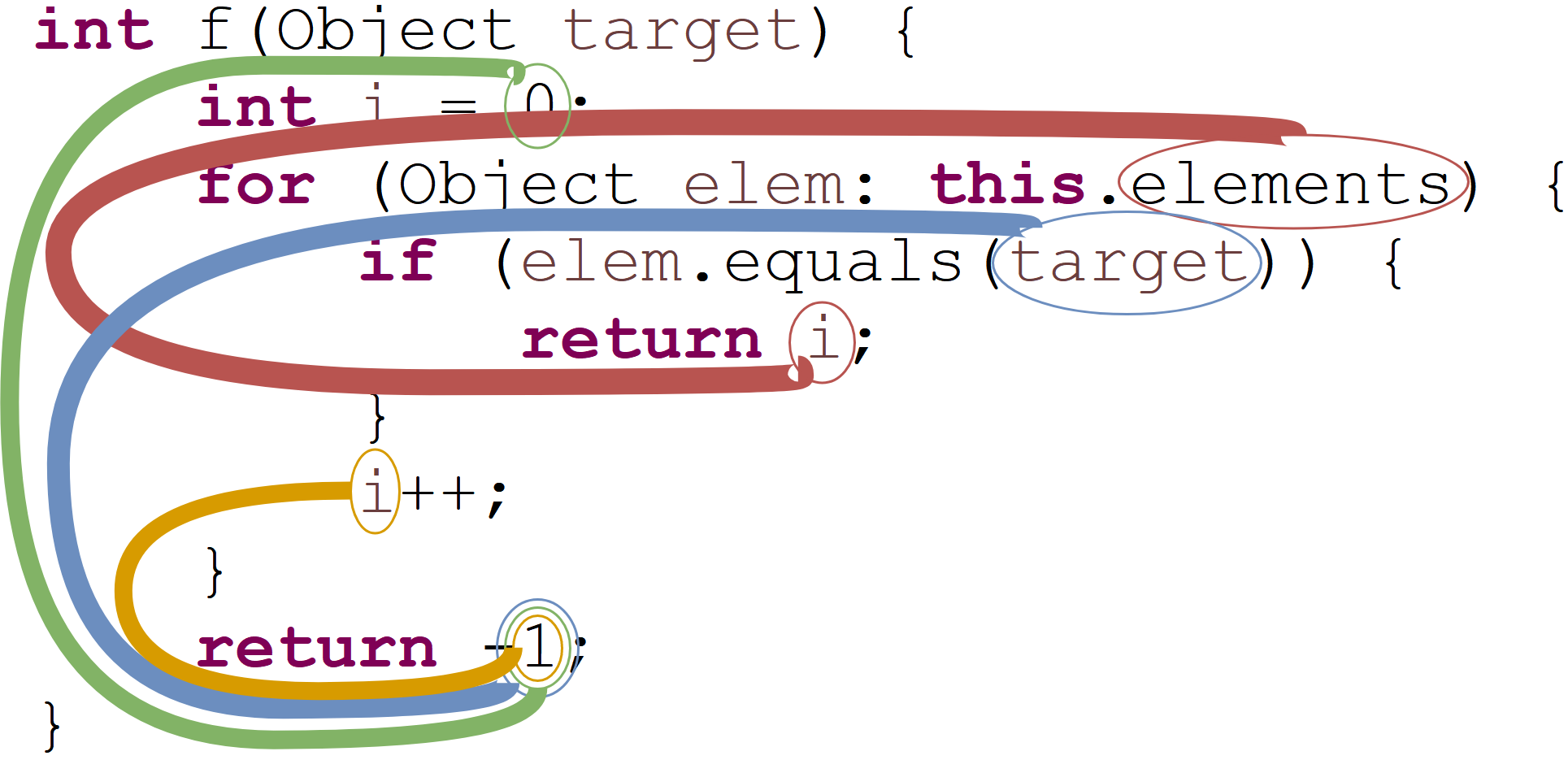}
\caption{}
\label{overview-c}
\vspace{1mm}
\end{subfigure}
\\

\begin{subfigure}[t]{0.28\textwidth}
\hspace{0.1in}
\scriptsize
\begin{tabular}{llr}
{\ul Predictions:} & &   \\
\textbf{contains} & \progressbar{0.9093}& $90.93\%$ \\
\textbf{matches} & \progressbar{0.0354} & $3.54\%$ \\
\textbf{canHandle} & \progressbar{0.0115} & $1.15\%$\\
\textbf{equals}& \progressbar{0.0087} & $0.87\%$\\
\textbf{containsExact} & \progressbar{0.0077} & $0.77\%$
\end{tabular}
\end{subfigure}
\rulesep
\begin{subfigure}[t]{0.32\textwidth}
\hspace{0.01in}
\scriptsize
\begin{tabular}{llr}
{\ul Predictions} & &   \\
\textbf{get}  &\progressbar{0.3109}& $31.09\%$ \\
\textbf{getProperty} & \progressbar{0.2025}& $20.25\%$ \\
\textbf{getValue}  &\progressbar{0.1434}& $14.34\%$ \\
\textbf{getElement} & \progressbar{0.14}& $14.00\%$ \\
\textbf{getObject} &\progressbar{0.0605}& $6.05\%$
\end{tabular}
\end{subfigure}
\rulesep
\begin{subfigure}[t]{0.3\textwidth}
\hspace{0.01in}
\scriptsize
\begin{tabular}{llr}
{\ul Predictions}  & &  \\
\textbf{indexOf} &\progressbar{0.9665}& $96.65\%$ \\
\textbf{getIndex} &\progressbar{0.0224}& $2.24\%$ \\
\textbf{findIndex} &\progressbar{0.0033}& $0.33\%$ \\
\textbf{indexOfNull} &\progressbar{0.0020}& $0.20\%$ \\
\textbf{getInstructionIndex}&\progressbar{0.0013}& $0.13\%$
\end{tabular}
\end{subfigure}

\end{tabular}

\end{minipage}
\caption{An example for three methods that 
albeit having
have a similar syntactic structure can be easily distinguished by our model; our model successfully captures the subtle differences between them and manages to predict meaningful names. Each method portrays the top-4 paths that were given the most attention by the model. The widths of the colored paths are proportional to the attention that each path was given.}
\label{overview-all}
\end{figure*}

In this section we demonstrate how our model assigns different vectors to similar snippets of code, \emph{in a way that captures the subtle differences between them}. The vectors are useful for making a prediction about each snippet, even though none of these snippets has been exactly observed in the training data.

The main idea of our approach is to extract syntactic paths from within a code snippet, represent them as a bag of distributed vector representations, and use an attention mechanism to compute a learned weighted average of the path vectors in order to produce a single \emph{code vector}. Finally, this code vector can be used for various tasks, such as to predict a likely name for the whole snippet.

\subsection{Motivating Example}

Since method names are usually descriptive and accurate labels for code snippets, we demonstrate our approach for the task of learning code vectors for method bodies, and predicting the method name given the body. In general, the same approach can be applied to any snippet of code that has a corresponding label.

Consider the three Java methods of \Cref{overview-all}. These methods share a similar syntactic structure: they all
\begin{inparaenum}[(i)]
\item have a single parameter named \scode{target}
\item iterate over a field named \scode{elements} and
\item have an \scode{if} condition inside the loop body.
\end{inparaenum}
The main differences are that the method of Fig.~\ref{overview-a} returns \scode{true} when \scode{elements} \emph{contains} \scode{target} and \scode{false} otherwise; the method of Fig.~\ref{overview-b} returns the element from \scode{elements} which \scode{target} \emph{equals to its \scode{hashCode}}; and the method of Fig.~\ref{overview-c} returns the \emph{index of} \scode{target} in \scode{elements}. Despite their shared characteristics, our model captures the subtle differences and predicts the descriptive method names: \scode{contains}, \scode{get}, and \scode{indexOf} respectively.

\begin{figure}
\centering
\includegraphics[width=5in]{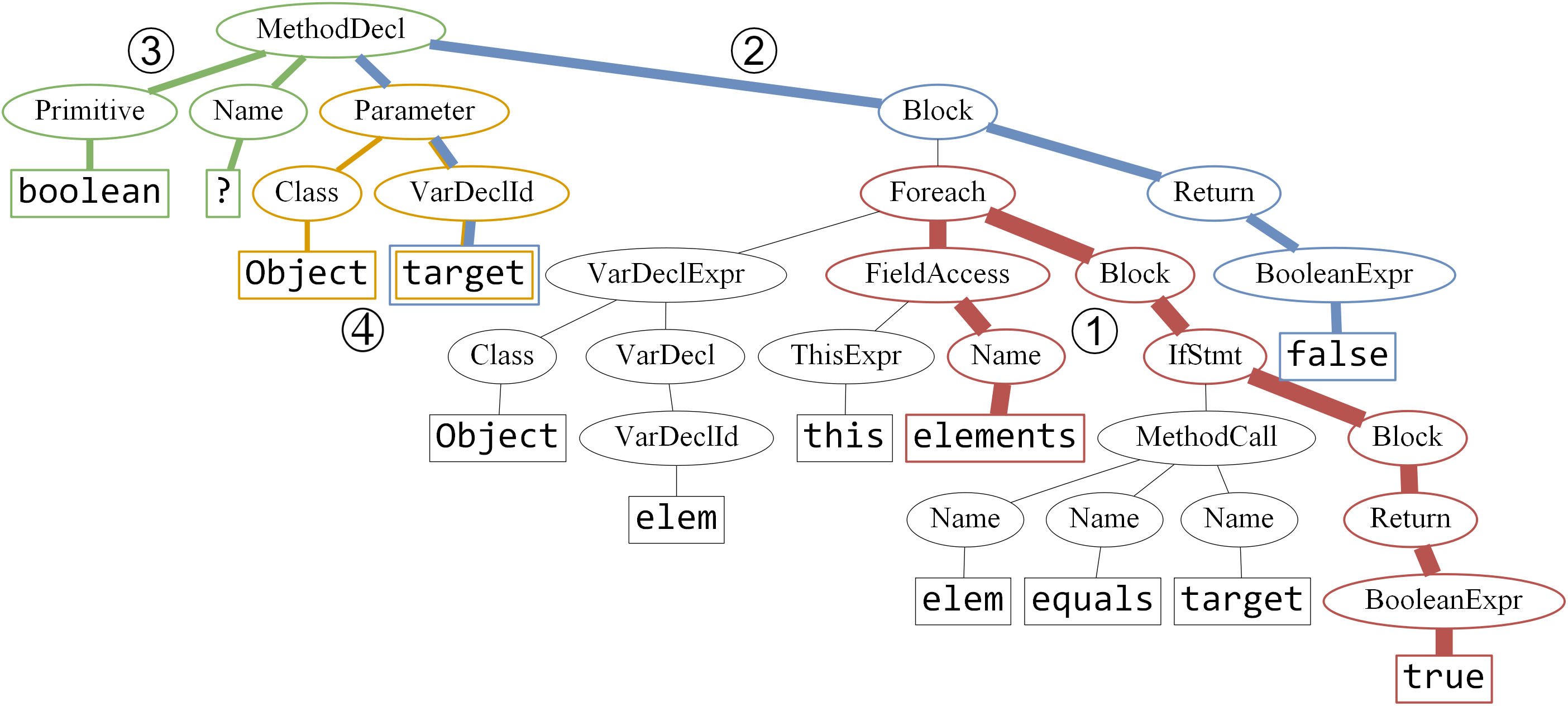}
\caption{The top-4 attended paths of \Cref{overview-a}, as were learned by the model, shown on the AST of the same  snippet. The width of each colored path is proportional to the attention it was given (\textbf{\textcolor{Maroon}{red}} \raisebox{.5pt}{\textcircled{\raisebox{-.9pt} {1}}}: $\textbf{0.23}$, \textbf{\textcolor{NavyBlue}{blue}} \raisebox{.5pt}{\textcircled{\raisebox{-.9pt} {2}}}: $\textbf{0.14}$, \textbf{\textcolor{path-green}{green}} \raisebox{.5pt}{\textcircled{\raisebox{-.9pt} {3}}}: $\textbf{0.09}$, \textbf{\textcolor{path-orange}{orange}} \raisebox{.5pt}{\textcircled{\raisebox{-.9pt} {4}}}: $\textbf{0.07}$).}
\label{contains-ast}
\end{figure} 

\paragraph{Path extraction}
First, each query method in the training corpus is parsed to construct an AST. Then, the AST is traversed and syntactic paths between AST leaves are extracted. Each path is represented as a sequence of AST nodes, linked by up and down arrows, which symbolize the up or down link between adjacent nodes in the tree. The path composition is kept with the values of the AST leaves it is connecting, as a tuple which we refer to as a \emph{path-context}. These terms are defined formally in \Cref{Background}. \Cref{contains-ast} portrays the top-four path-contexts that were given the most attention by the model, on the AST of the method from \Cref{overview-a}, such that the width of each path is proportional to the attention it was given by the model during this prediction.

\paragraph{Distributed representation of contexts}
Each of the path and leaf-values of a path-context is mapped to its corresponding real-valued vector representation, or its \emph{embedding}. Then, the three vectors of each context are concatenated to a single vector that represents that path-context. During training, the values of the embeddings are learned jointly with the attention parameter and the rest of the network parameters.

\paragraph{Path-attention network}
The Path-Attention network aggregates multiple path-contexts embeddings into a single vector that represents the whole method body. Attention is the mechanism that learns to score each path-context, such that higher attention is reflected in a higher score.
These multiple embeddings are aggregated using the attention scores into a single \emph{code vector}. The network then predicts the probability for each target method name given the code vector. The network architecture is described in~\Cref{Model}.

\paragraph{Path-attention interpretation}
While it is usually difficult or impossible to interpret specific values of vector components in neural networks, it is possible and interesting to observe the attention scores that each path-context was given by the network.
Each code snippet in \Cref{overview-all} and \Cref{contains-ast} highlights the top-four path-contexts that were given the most weight (attention) by the model in each example.
The widths of the paths are proportional to the attention score that each of these path-contexts was given.
The model has learned how much weight to give every possible path on its own, as part of training on millions of examples.
For example, it can be seen in \Cref{contains-ast} that the \textbf{\textcolor{Maroon}{red}} \raisebox{.5pt}{\textcircled{\raisebox{-.9pt} {1}}} path-context, which spans from the field \scode{elements} to the return value \scode{true} was given the highest attention. For comparison, the \textbf{\textcolor{NavyBlue}{blue}} \raisebox{.5pt}{\textcircled{\raisebox{-.9pt} {2}}} path-context, which spans from the parameter \scode{target} to the return value \scode{false} was given a lower attention.

Consider the \textbf{\textcolor{Maroon}{red}} \raisebox{.5pt}{\textcircled{\raisebox{-.9pt} {1}}} path-context of \Cref{overview-a} and \Cref{contains-ast}. As we explain in \Cref{Background}, this path is represented as:
{\footnotesize
\begin{equation}
(\scode{elements},
\textrm{Name}{\uparrow}\textrm{FieldAccess}{\uparrow}\textrm{Foreach}{\downarrow}\textrm{Block}{\downarrow}\textrm{IfStmt}{\downarrow}\textrm{Block}{\downarrow}\textrm{Return}{\downarrow}\textrm{BooleanExpr}, \scode{true}) \nonumber
\end{equation}
}%

Inspecting this path node-by-node reveals that this single path captures the main functionality of the method: the method iterates over a field called \scode{elements}, and for each of its values it checks an \scode{if} condition; if the condition is true, the method returns \scode{true}. Since we use soft-attention, the final prediction takes into account other paths as well, such as paths that describe the \scode{if} condition itself, but it can be understood why the model gave this path the highest attention.

\Cref{overview-all} also shows the top-5 suggestions from the model for each method. As can be seen in all of the three examples, in many cases most of the top suggestions are very similar to each other and all of them are descriptive regarding the method. Observing the top-5 suggestions in \Cref{overview-a} shows that two of them (\scode{contains} and \scode{containsExact}) are very accurate, but it can also be imagined how a method called \scode{matches} would share similar characteristics: a method called \scode{matches} is also likely to have an \scode{if} condition inside a \scode{for} loop, and to return \scode{true} if the condition is true.

Another interesting observation is that the \textbf{\textcolor{path-orange}{orange}} \raisebox{.5pt}{\textcircled{\raisebox{-.9pt} {4}}} path-context of \Cref{overview-a} which spans from \scode{Object} to \scode{target} was given a lower attention than other path-contexts in the same method, but \emph{higher attention than the same path-context in \Cref{overview-c}}. This demonstrates how attention is not constant, but given with respect to the other path-contexts in the code.

\paragraph{Comparison with n-grams}
The method in \Cref{overview-a} shows the four path-contexts that were given the most attention during the prediction of the method name \scode{contains}. Out of them, the \textbf{\textcolor{path-orange}{orange}} \raisebox{.5pt}{\textcircled{\raisebox{-.9pt} {4}}} path-context spans between two consecutive tokens: \scode{Object} and \scode{target}. This might create the (false) impression that representing this method as a bag-of-bigrams could be as expressive as syntactic paths. However, as can be seen in \Cref{contains-ast}, the \textbf{\textcolor{path-orange}{orange}} \raisebox{.5pt}{\textcircled{\raisebox{-.9pt} {4}}} path goes through an AST node of type $Parameter$, which uniquely distinguishes it from, for example, a local variable declaration of the same name and type. In contrast, a bigram model will represent the expression \scode{Object target} equally whether \scode{target} is a method parameter or a local variable. This shows that a model using a syntactic representation of a code snippet can distinguish between two snippets of code that other representations cannot, and by aggregating all the contexts using attention, all these subtle differences contribute to the prediction.

\paragraph{Key aspects}
The illustrated examples highlight several key aspects of our approach:
\begin{itemize}
\item A code snippet can be efficiently represented as a bag of path-contexts.
\item Using a single context is not enough to make an accurate prediction. An attention-based neural network can identify the importance of multiple path-contexts, and aggregate them accordingly to make a prediction.
\item Subtle differences between code snippets are easily distinguished by our model, even if the code snippets have a similar syntactic structure and share many common tokens and n-grams.
\item Large corpus, cross-project prediction of method names is possible using this model.
\item Although our model is based on a neural network, the model is human-interpretable and provides interesting observations.
\end{itemize} 
\section{Background - Representing Code using AST Paths}\label{Background}

In this section, we briefly describe the representation of a code snippet as a set of syntactic paths in its abstract syntax tree (AST). This representation is based on the general-purpose approach for representing program elements by \citet{pigeon-pldi}. The main difference in this definition is that we define this representation to handle \emph{whole snippets of code}, rather than a single program element (such as a single variable), and use it as input to our path-attention neural network.

We start by defining an AST, a path and a path-context.

\begin{definition}[Abstract Syntax Tree]
An Abstract Syntax Tree (AST) for a code snippet $\mathcal{C}$ is a tuple $\langle N, T, X, s, \delta, \phi \rangle$ where $N$ is a set of nonterminal nodes, $T$ is a set of terminal nodes, $X$ is a set of values, $s\in N$ is the root node, $\delta: N\rightarrow \left(N\cup T\right)^*$ is a function that maps a nonterminal node to a list of its children, and $\phi:T\rightarrow X$ is a function that maps a terminal node to an associated value. Every node except the root appears exactly once in all the lists of children.
\end{definition}


Next, we define AST paths. For convenience, in the rest of this section we assume that all definitions refer to a single AST $ \langle N, T, X, s, \delta, \phi \rangle$.

An AST path is a path between nodes in the AST, starting from one terminal, ending in another terminal, passing through an intermediate nonterminal in the path which is a common ancestor of both terminals. More formally:

\begin{definition}[AST path]
An AST-path of length $k$ is a sequence of the form: $n_{1}d_{1}...n_{k}d_{k}n_{k+1}$, where $n_{1},n_{k+1}\in T$ are terminals, for $i\in \left[2..k\right]$: $n_{i}\in N$ are nonterminals and for $i\in \left[1..k\right]$:  $d_{i} \in \{ \uparrow, \downarrow\}$ are movement directions (either up or down in the tree).
If $d_{i}=\uparrow$, then: $n_{i}\in\delta\left(n_{i+1}\right)$; if $d_{i}=\downarrow$, then: $n_{i+1}\in\delta\left(n_{i}\right)$. For an AST-path $p$, we use $start\left(p\right)$ to denote $n_{1}$ - the starting terminal of $p$; and $end\left(p\right)$ to denote $n_{k+1}$ - its final terminal.
\end{definition}

Using this definition we define a \emph{path-context} as a tuple of an AST path and the values associated with its terminals:

\begin{definition}[Path-context]
Given an AST Path $p$, its path-context is a triplet $\langle x_{s},p,x_{t} \rangle$ where $x_{s}\!=\! \phi\left(start\left(p\right)\right)$ and $x_{t}\!=\!\phi\left(end\left(p\right)\right)$ are the values associated with the start and end terminals of $p$.
\end{definition}

That is, a path-context describes two actual tokens with the syntactic path between them.



\begin{example}
A possible path-context that represents the statement: ``\scode{x = 7;}'' would be:
\begin{equation*}
\small
\langle \scode{x},\left(NameExpr\uparrow AssignExpr\downarrow IntegerLiteralExpr\right),\scode{7} \rangle
\end{equation*}
 	
\end{example}

Practically, to limit the size of the training data and reduce sparsity, it is possible to limit the paths by different aspects. Following earlier works, we limit the paths by maximum \emph{length} --- the maximal value of $k$, and limit the maximum \emph{width} --- the maximal difference in child index between two child nodes of the same intermediate node. These values are determined empirically as hyperparameters of our model.

\section{Model}\label{Model}

In this section we describe our model in detail. \Cref{Representation} describes the way the input source code is represented; \Cref{NeuralNetwork} describes the architecture of the neural network; \secref{Training} describes the training process, and \secref{Prediction} describes the way the trained model is used for prediction. Finally \secref{Design} discusses some of the model design choices, and compares the architecture to prior art.

\paragraph{High-level view} At a high-level, the key point is that a code snippet is composed of a bag of contexts, and each context is represented by a vector that its values are learned. The values of this vector capture two distinct goals:
\begin{inparaenum}[(i)]
\item the semantic meaning of this context, and
\item the amount of attention this context should get.
\end{inparaenum}

The problem is as follows: given an arbitrarily large number of context vectors, we need to aggregate them into a single vector. Two trivial approaches would be to learn the most important one of them, or to use them all by vector-averaging them. These alternatives will be discussed in \Cref{alternative}, and the results of implementing these two alternatives are shown in \Cref{soft-hard} (``hard-attention'' and ``no-attention'') to yield poor results.

The main understanding in this work is that \emph{all} context vectors need to be used, but the model should be let to learn how much focus to give each vector. This is done by learning how to average context vectors in a weighted manner. The weighted average is obtained by weighting each vector by a factor of its dot product with another global attention vector. The vector of each context and the global attention vector are trained and learned \emph{simultaneously} using the standard neural approach of backpropagation. Once trained, the neural network is simply a pure mathematical function, which uses algebraic operators to output a code vector given a set of contexts.

\subsection{Code as a Bag of Path-Contexts}\label{Representation}

Our path-attention model receives as input a code snippet in some programming language and a parser for that language.


\paragraph{Representing a snippet of code}
We denote by $Rep$ the representation function (also known as a feature function) which transforms a code snippet into a mathematical object that can be used in a learning model. Given a code snippet $\mathcal{C}$ and its AST $\langle N, T, X, s, \delta, \phi \rangle$,
we denote by $TPairs$ the set of all pairs of AST terminal nodes (excluding pairs that contain a node and itself):
\begin{equation*}
TPairs\left( C\right)=\left\{\left( term_i,term_j\right) |term_i,term_j\in termNodes\left(C\right)\land i\neq j\right\}
\end{equation*}
where $termNodes$ is a mapping between a code snippet and the set of terminal nodes in its AST. We represent $\mathcal{C}$ as the set of path-contexts that can be derived from it:

\[
  Rep\left(\mathcal{C}\right) =\left\lbrace \left(x_s,p,x_t\right) \;\middle|\;
  \begin{tabular}{@{}l@{}}
    $\exists (term_s,term_t)\in TPairs\left(\mathcal{C}\right): $\\
    $x_s=\phi\left(term_s\right)\land x_t=\phi\left(term_t\right)$ \\
    $\land\, start(p) = term_s \land end(p) = term_t$
   \end{tabular}
  \right\rbrace
\]

that is, $\mathcal{C}$ is represented as the set of triplets $\langle x_s,p,x_t \rangle$ such that $x_s$ and $x_t$ are values of AST terminals, and $p$ is the AST path that connects them. For example, the representation of the code snippet from \Cref{overview-a} contains, among others, the four AST paths of \Cref{contains-ast}.

\subsection{Path-Attention Model}\label{NeuralNetwork}

\begin{figure}
\centering
\includegraphics[width=4in]{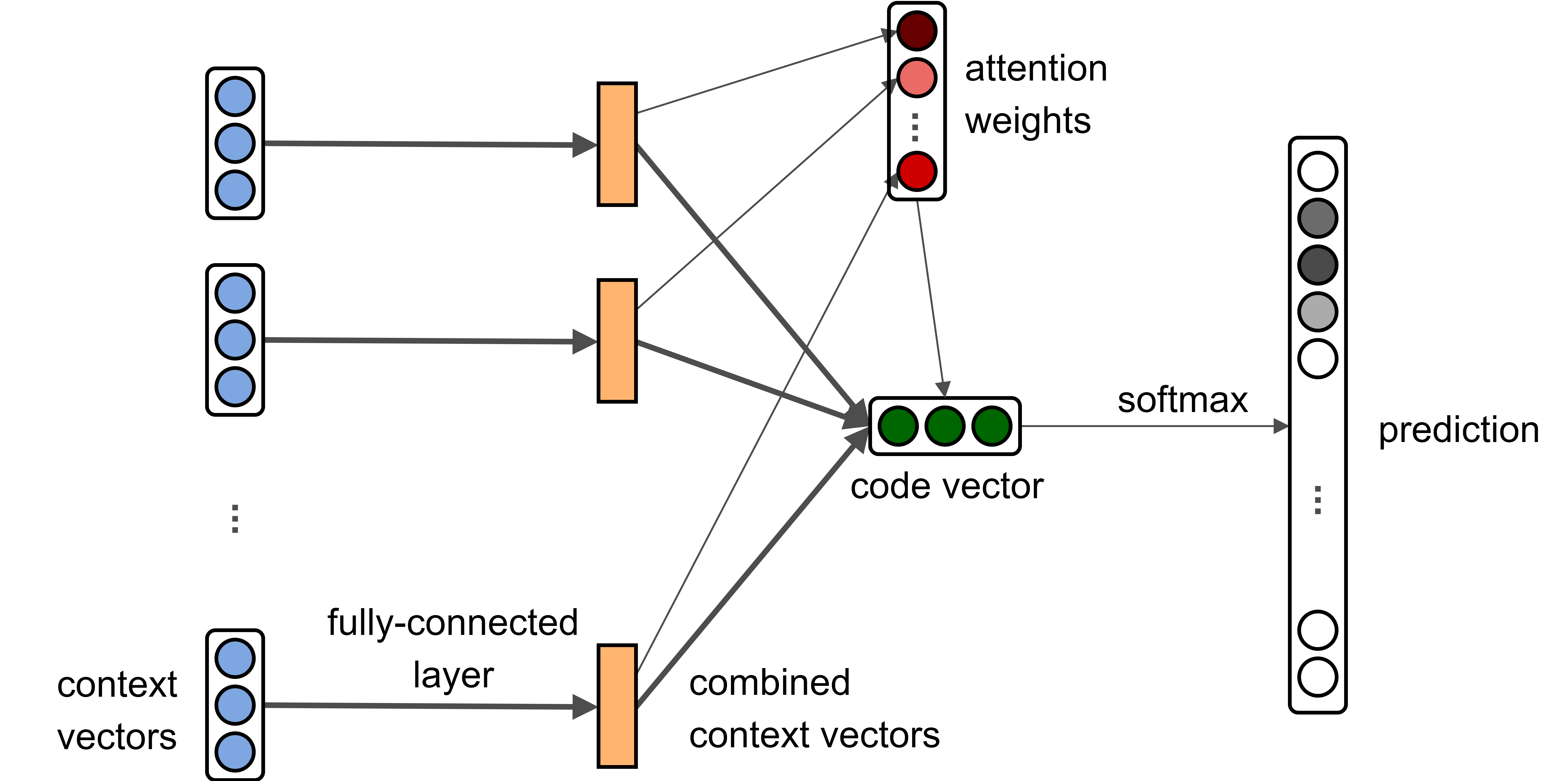}
\caption{The architecture of our path-attention network. A \emph{full-connected layer} learns to combine embeddings of each path-contexts with itself; attention weights are learned using the combined context vectors, and used to compute a \emph{code vector}. The code vector is used to predicts the label.}
\figlabel{network}
\end{figure}

Overall, the model learns the following components: embeddings for paths and names (matrices $path\_vocab$ and $value\_vocab$), a fully connected layer (matrix $W$), attention vector ($\boldsymbol{a}$), and embeddings for the tags ($tags\_vocab$).
We describe our model from from left-to-right (\figref{network}).
We define two embedding vocabularies: $value\_vocab$ and $path\_vocab$, which are matrices in which every row corresponds to an embedding associated with a certain object:
\begin{align*}
value\_vocab &\in \mathbb{R}^{\left|X\right|\times d} \\
path\_vocab &\in \mathbb{R}^{\left|P\right|\times d}
\end{align*}
where as before, $X$ is the set of values of AST terminals that were observed during training, and $P$ is the set of AST paths. Looking up an embedding is simply picking the appropriate row of the matrix. For example, if we consider \Cref{overview-a} again, $value\_vocab$ contains rows for each token value such as \scode{boolean}, \scode{target} and \scode{Object}. $path\_vocab$ contains rows which are mapped to each of the AST paths of \Cref{contains-ast} (without the token values), such as the \textbf{\textcolor{Maroon}{red}} \raisebox{.5pt}{\textcircled{\raisebox{-.9pt} {1}}} path:
$Name\uparrow FieldAccess\uparrow Foreach\downarrow Block\downarrow IfStmt\downarrow Block\downarrow Return\downarrow BooleanExpr$.
The values of these matrices are initialized randomly and are learned simultaneously with the network during training.

The width of the matrix $W$ is the embeddings size $d\in \mathbb{N}$ -- the dimensionality hyperparameter. $d$ is determined empirically, limited by the training time, model complexity and the GPU memory, and typically ranges between 100-500. For convenience, we refer to the embeddings of both the paths and the values as vectors of the same size $d$, but in general they can be of different sizes.

A bag of path-contexts $\mathcal{B}=\left\{b_1,...,b_n\right\}$ that were extracted from a given code snippet is fed into the network.
Let $b_{i}=\langle x_{s},p_j,x_{t} \rangle$ be one of these path-contexts, such that $\left\{x_{s},x_{t}\right\}\in X$ are values of terminals and $p_j \in P$ is their connecting path. Each component of a path-context is looked-up and mapped to its corresponding embedding. The three embeddings of each path-context are concatenated to a single \emph{context vector}: $\boldsymbol{c_i}\in\mathbb{R}^{3d}$ that represents that path-context:
\begin{equation}
\boldsymbol{c_{i}} = 
embedding\left(\langle x_{s},p_j,x_{t} \rangle\right)= \\
\left[\begin{array}{l}
value\_vocab_{s} \,;\, path\_vocab_{j} \,;\, value\_vocab_{t}
\end{array}
\right] \in \mathbb{R}^{3d} \\
\end{equation}

For example, for the \textbf{\textcolor{Maroon}{red}} \raisebox{.5pt}{\textcircled{\raisebox{-.9pt} {1}}} path-context from \Cref{contains-ast}, its context vector would be the concatenation of the vectors of \scode{elements}, the \textbf{\textcolor{Maroon}{red}} \raisebox{.5pt}{\textcircled{\raisebox{-.9pt} {1}}} path, and \scode{true}.

\paragraph{Fully connected layer}
Since every context vector $\boldsymbol{c_i}$ is formed by a concatenation of three independent vectors, a fully connected layer learns to \emph{combine} its components. This is done separately for each context vector, using the same learned combination function. This allows the model to give a different attention to every \emph{combination} of paths and values. This combination allows the model the expressivity of giving a certain path a high attention when observed with certain values, and a low attention when the exact same path is observed with other values.

Here, $\boldsymbol{\tilde{c}_i}$ is the output of the fully connected layer, which we refer to as a \emph{combined context vector}, computed for a path-context $b_i$. The computation of this layer can be described simply as:
\begin{equation*}
\boldsymbol{\tilde{c}_i} = tanh\left(W \cdot \boldsymbol{c_i}\right)
\end{equation*}
where $W\in\mathbb{R}^{d\times 3d}$ is a learned weights matrix and $tanh$ is the hyperbolic tangent function.
The height of the weights matrix $W$ determines the size of $\boldsymbol{\tilde{c}_i}$, and for convenience is the same size $d$ as before, but in general it can be of a different size to determines the size of the \emph{target} vector.
$tanh$ is the hyperbolic tangent element-wise function, a commonly used monotonic nonlinear activation function which outputs values that range $\left(-1,1\right)$, which increases the expressiveness of the model.

That is, the fully connected layer ``compresses'' a context vector of size $3d$ into a combined context vector of size $d$ by multiplying it with a weights matrix, and then applies the $tanh$ function to each element of the vector separately.

\paragraph{Aggregating multiple contexts into a single vector representation with attention}
The attention mechanism computes a weighted average over the combined context vectors, and its main job is to compute a scalar weight to each of them.
An attention vector $\boldsymbol{a}\in\mathbb{R}^{d}$ is initialized randomly and learned simultaneously with the network. Given the combined context vectors: $\left\{\boldsymbol{\tilde{c}_1},...,\boldsymbol{\tilde{c}_n}\right\}$,
the attention weight $\alpha_{i}$ of each $\boldsymbol{\tilde{c}_i}$ is computed as the normalized inner product between the combined context vector and the global attention vector $\boldsymbol{a}$:
\begin{equation*}
\text{attention weight }\alpha_i = \frac{\exp(\boldsymbol{\tilde{c}_{i}}^{T}\cdot \boldsymbol{a})}{\sum_{j=1}^{n}\exp(\boldsymbol{\tilde{c}_{j}}^{T}\cdot \boldsymbol{a})}
\end{equation*}

The exponents in the equations are used only to make the attention weights positive, and they are divided by their sum to have a sum of $1$, as a standard softmax function.

The aggregated code vector $\boldsymbol{\upsilon}\in \mathbb{R}^{d}$, which represents the whole code snippet, is a linear combination of the combined context vectors $\left\{\boldsymbol{\tilde{c}_1},...,\boldsymbol{\tilde{c}_n}\right\}$ factored by their attention weights:
\begin{equation}
\text{code vector } \boldsymbol{\upsilon} = \sum_{i=1}^{n}\alpha_i \cdot \boldsymbol{\tilde{c}_i}
\label{codevector}
\end{equation}

that is, the attention weights are non-negative and their sum is $1$, and they are used as the factors of the combined context vectors $\boldsymbol{\tilde{c}_i}$. Thus, attention can be viewed as a weighted average, where the weights are learned and calculated with respect to the other members in the bag of path-contexts.

\paragraph{Prediction}
Prediction of the tag is performed using the code vector. We define a tags vocabulary which is learned as part of training:
\begin{equation*}
tags\_vocab\in \mathbb{R}^{\left|Y\right|\times d}
\end{equation*}

where $Y$ is the set of tag values found in the training corpus. Similarly as before, the embedding of $tag_i$ is row $i$ of $tags\_vocab$.
For example, looking at \Cref{overview-a} again, $tags\_vocab$ contains rows for each of \scode{contains}, \scode{matches} and \scode{canHandle}.
The predicted distribution of the model $q\left (y\right)$ is computed as the (softmax-normalized) dot product between the code vector $\boldsymbol{\upsilon}$ and each of the tag embeddings:
\begin{equation*}
for\, y_{i}\in Y:\, q\left(y_i\right)=\frac{\exp(\boldsymbol{\upsilon}^T\cdot tags\_vocab_{i})}{\sum_{y_{j}\in Y}\exp(\boldsymbol{\upsilon}^T\cdot tags\_vocab_{j})}
\end{equation*}

that is, the probability that a specific tag $y_i$ should be assigned to the given code snippet $\mathcal{C}$ is the normalized dot product between the vector of $y_i$ and the code vector $\boldsymbol{\upsilon}$.

\subsection{Training}\seclabel{Training}
For training the network we use cross-entropy loss~\cite{rubinstein1999cross, rubinstein2001combinatorial} between the predicted distribution $q$ and the ``true'' distribution $p$. Since $p$ is a distribution that assigns a value of $1$ to the actual tag in the training example and $0$ otherwise, the cross-entropy loss
for a single example is equivalent to the negative log-likelihood of the true label, and
can be expressed as:
\begin{equation*}
\mathcal{H}\left(p||q\right)=-\sum_{y\in Y}p\left(y\right)log\,q\left(y\right)=-log\,q\left(y_{true}\right)
\end{equation*}

where $y_{true}$ is the actual tag that was seen in the example. That is, the loss is the negative logarithm of  $q\left(y_{true}\right)$, the probability that the model assigns to $y_{true}$. As $q\left(y_{true}\right)$ tends to $1$, the loss approaches zero. The further $q\left(y_{true}\right)$ goes below $1$, the greater the loss becomes. Thus, minimizing this loss is equivalent to maximizing the log-likelihood that the model assigns to the true labels $y_{true}$.

Training the network is performed using any gradient descent based algorithm, and the standard approach of backpropagating the training error through each of the learned parameters (i.e., deriving the loss with respect to each of the learned parameters and updating the learned parameter's value by a small ``step'' towards the direction that minimizes the loss).

\subsection{Using the trained network}\seclabel{Prediction}
A trained network can be used for two main purposes:
\begin{inparaenum}[(i)]
\item Use the code vector $\boldsymbol{\upsilon}$ itself in a down-stream task, and
\item Use the network to predict tags for new, unseen code.
\end{inparaenum}

\paragraph{Using the code vector} An unseen code can be fed into the trained network exactly the same as in the training step, up to the computation of the code vector (Eq. \eqref{codevector}). This code embedding can now be used in another deep learning pipeline for various tasks such as finding similar programs, code search, refactoring suggestion, and code summarization.

\paragraph{Predicting tags and names} the network can also be used to predict tags and names for unseen code. In this case we also compute the code vector $\boldsymbol{\upsilon}$ using the weights and parameters that were learned during training, and prediction is done by finding the closest target tag:
\begin{equation*}
prediction\left(\mathcal{C}\right)=
argmax_{\mathcal{L}}P(\mathcal{L}|\mathcal{C})=
argmax_{\mathcal{L}}\left\{q_{\upsilon_\mathcal{C}}(y_\mathcal{L})\right\}
\end{equation*}

where $q_{\upsilon_\mathcal{C}}$ is the predicted distribution of the model, given the code vector $\boldsymbol{\upsilon_\mathcal{C}}$.

\paragraph{Scenario-dependant variants} For simplicity, we describe a network that predicts a single label, but the same architecture can be adapted for slightly different scenarios. For example, in a multi-tagging scenario \cite{tsoumakas2006multilabel}, each code snippet contains multiple true tags as in StackOverflow questions. Another example is predicting a sequence of target words such as method documentation. In the latter case, the attention vector should be used to re-compute the attention weights after each predicted token, given the previous prediction, as commonly done in neural machine translation \cite{bahdanau14, luong15}.

\subsection{Design Decisions}\seclabel{Design}
\paragraph{Bag of contexts}
We represent a snippet of code as an unordered  bag of path-contexts. This choice reflects our hypothesis that the \emph{existence} of path-contexts in a method body is more significant than their internal location or order.

An alternative representation is to sort path-contexts according to a predefined order (e.g., order of their occurrence). However, unlike natural language, there is no predetermined location in a method where the main attention should be focused. An important path-context can appear anywhere in a method body (and span throughout the method body).

\paragraph{Working with syntactic-only context}
The main contribution of this work is its ability to aggregate multiple contexts into a fixed-length vector in a weighted manner, and use the vector to make a prediction. In general, our proposed model is not bound to any specific representation of the input program, and can be applied in a similar way to a ``bag of contexts'' in which the contexts are designed for a specific task, or contexts that were produced using semantic analysis. Specifically, we chose to use a syntactic representation that is similar to \citet{pigeon-pldi} because it was shown to be useful as a representation for modelling programming languages in machine learning models, and more expressive than n-grams and manually-designed features.

An alternative approach is to include semantic relations as context. Such an approach was performed by \citet{allamanis2017learning} who presented a Gated Graph Neural Network, in which program elements are graph nodes and semantic relations such as \scode{ComputedFrom} and \scode{LastWrite} are edges in the graph. In their work, these semantic relations were chosen and implemented for specific programming language and tasks.
In our work, we wished to explore \emph{how far can a syntactic-only approach go}. Using semantic knowledge has many advantages and potentially contains information that is not clearly expressed in a syntactic-only observation, but comes at a cost: (i) an expert is required to choose and design the semantic analyses; (ii) generalizing to new languages is much more difficult, as the semantic analyses need to be implemented differently for every language; and (iii) the designed analyses might not easily generalize to other tasks.
In contrast, in our syntactic approach (i) no expert knowledge of the language nor manual feature designing is required; (ii) generalizing to other languages is performed by simply replacing the parser and extracting paths from the new language's AST using the same traversal algorithm; and (iii) the same syntactic paths generalize surprisingly well to other tasks (as was shown by \citet{pigeon-pldi}).

\paragraph{Large corpus, simple model}
Similarly to the approach of~\citet{mikolovEfficient2013} for word representations, we found that it is more efficient to use a simpler model with a large amount of data, rather than a complex model and a small corpus.

Some previous works decomposed the target predictions. \citet{conv16, allamanis2015} decomposed method names into smaller ``sub-tokens'' and used the continuous prediction approach to compose a full name. \citet{codenn16} decomposed StackOverflow titles to single words and predicted them word-by-word. In theory, this approach could be used to predict new compositions of names that were not observed in the training corpus, referred to as neologisms \cite{allamanis2015}. However, when scaling to millions of examples this approach might become cumbersome and fail to train well due to hardware and time limitations. As shown in \Cref{quantitative}, our model yields significantly better results than previous models that used this approach.


Another disadvantage of subtoken-by-subtoken learning is that it requires a time consuming beam-search during prediction. This results in an \emph{orders of magnitude slower prediction rate} (the number of predictions that the model is able to make per second). An empirical comparison of the prediction rate of our model and the models of \citet{conv16, codenn16}, shows that our model achieves roughly $200$ times higher prediction rate than \citet{codenn16} and $10,000$ times higher than \citet{conv16} (\Cref{quantitative}).

\paragraph{OoV prediction}
The main potential advantage of the models of \citet{conv16} and \citet{codenn16} over our model is the subtoken-by-subtoken prediction, which allows them to predict a neologism, and the copy mechanism used by \citet{conv16} which allows it to use out-of-vocabulary (OoV) words in the prediction.

An analysis of our test data shows that the top-10 most frequent method names, such as \scode{toString}, \scode{hashCode} and \scode{equals}, which are typically easy to predict, appear in less than $6\%$ of the test examples. The $13\%$ least occurring names are rare names, which did not appear as whole in the training data, and are difficult or impossible to predict exactly even with a neologism or copy mechanism, such as: \scode{imageFormatExceptionShouldProduceNotSuccessOperationResultWithMessage}. 
Therefore, our goal is to maximize our efforts on the remaining of the examples.

Even though the upper bound of accuracy of models which incorporate neologism or copy mechanisms is hypothetically higher than ours, the actual contribution of these abilities is minor. Empirically, when trained and evaluated on the same corpora as our model, only less than $3\%$ of the predictions of each of these baselines were actually neologism or OoV. Further, out of all the cases that the baseline suggested a neologism or OoV, \emph{more predictions could have been exact-matches using an already-seen target name, rather than composing a neologism or OoV}.

Although it is possible to incorporate these mechanisms in our model as well, we chose to predict complete names due to the high cost of training and prediction time and the relatively negligible contribution of these mechanisms.

\paragraph{Granularity of path decomposition}
An alternative approach could decompose the representation of a path to granularity of single nodes, and learn to represent a whole path node-by-node using a recurrent neural network (RNN).
This would possibly require less space, but will require more time to train and predict, as training of RNNs is usually more time consuming and not clearly better.

Further, a statistical analysis of our corpus shows that more than \emph{$95\%$ of the paths in the test set were already seen in the training set}.
Accordingly, in the trade-off between time and space we chose a little less expressive, more memory-consuming, but fast-to-train approach. This choice leads to results that are as $95\%$ as good as our final results in only $6$ hours of training, while significantly improving over previous works.
Despite our choice of time over space, training our model on millions of examples fits in the memory of common GPUs. 
\section{Distributed vs. Symbolic Representations}\label{Parameters}
We compare our model, which uses \emph{distributed representations}, with Conditional Random Fields (CRFs) as an example of a model that uses \emph{symbolic representations} (\cite{pigeon-pldi} and \cite{jsnice2015}). Distributed representations refer to representations of elements that are discrete in their nature (e.g. words and names) as vectors or matrices, such that the meaning of an element is distributed across multiple components. This contrasts with symbolic representations, where each element is uniquely represented with exactly one component \cite{allamanis2017survey}. Distributed representations have recently become extremely common in machine learning and NLP because they generalize better, while often requiring fewer parameters.

In general, CRFs can also use distributed representations \cite{durrett2015neural, artieres2010neural}, but for the purpose of this discussion, ``CRFs'' refers to a CRFs with symbolic representations as used by \citet{pigeon-pldi} and \citet{jsnice2015}.

\paragraph{Generalization ability}
Using CRFs for predicting program properties was found to be powerful \cite{jsnice2015, pigeon-pldi}, but limited to modeling only combinations of values that were seen in the training data.
In their works, in order to score the likelihood of a combination of values, the trained model keeps a scalar parameter for every
combination of three components that was observed in the training corpus: variable name, another identifier, and the relation between them. 
When an unseen combination is observed in test data, a model of this kind cannot generalize and evaluate its likelihood, even if each of the individual values was observed during training.

In contrast, the main advantage of distributed representations in this aspect is the ability to compute the likelihood of \emph{every} combination of observed values. Instead of keeping a parameter for every observed \emph{combination} of values, our model keeps a small constant number ($d$) of learned parameters for each atomic value, and use algebraic operations to compute the likelihood of their combination.

\paragraph{Trading polynomial complexity with linear} Using symbolic representations can be very costly in terms of the number of required parameters.
Using CRFs in our problem, which models the probability of a label given a bag of path-contexts, would require using \emph{ternary} factors, which require keeping a parameter for every observed combination of \emph{four} components: terminal value, path, another terminal value, and the target code label (a ternary factor which is determined by the path, with its three parameters). A CRF would thus have a space complexity of $O\left(\left|X\right|^2\cdot \left|P\right|\cdot \left|Y\right|\right)$, where $X$ is the set of terminal values, $P$ is the set of paths, and $Y$ is the set of labels.

In contrast, the number of parameters in our model is $O\left(d\cdot\left(\left|X\right|+\left|P\right|+\left|Y\right|\right)\right)$, where $d$ is a relatively small constants ($128$ in our final model) -- we keep a vector of size $d$ for every atomic terminal value or path, and use algebraic operations to compute the vector that represents the whole code snippet. Thus, distributed representations allow to \emph{trade the polynomial complexity with linear}. This is extremely important in these settings, because $\left|X\right|$, $\left|Y\right|$ and $\left|P\right|$ are in the orders of millions (the number of observed values, paths and labels). In fact, using distributed representations of symbols and relations in neural networks allows to keep \emph{less} parameters than CRFs, and at the same time compute a score to \emph{every} possible combination of observed values, paths and target labels, instead of only observed combinations.


Practically, we reproduced the experiments of \citet{pigeon-pldi} of modeling the task of predicting method names with CRFs using trenary factors. In addition to yielding an F1 score of $49.9$, which our model relatively improves by $17\%$, the CRF model required $+104\%$ more parameters, and about $10$ times more memory.

\section{Evaluation}\label{Evaluation}
The main contribution of our method is in its ability to aggregate an arbitrary sized snippet of code into a fixed-size vector in a way that captures its semantics.
Since Java methods are usually short, focused, have a single responsibility and a descriptive name, a natural benchmark of our approach would consider a method body as a code snippet, and use the produced code vector to predict the method name. Succeeding in this task would suggest that the code vector has indeed accurately captured the functionality and semantic role of the method.

Our evaluation aims to answer the following questions:
\begin{itemize}
\item How useful is our model in predicting method names, and how well does it measure in comparison to other recent approaches (\Cref{quantitative})?
\item What is the contribution of the attention mechanism to the model? How well would it perform using \emph{hard}-attention instead, or without attention at all (\Cref{alternative})?
\item What is the contribution of each of the path-context components to
    the model (\Cref{data_study})?
\item Is it actually able to predict names of complex methods, or only of trivial ones (\Cref{qualitative})?
\item What are the properties of the learned vectors? Which semantic patterns do they encode (\Cref{qualitative})?
\end{itemize}

\paragraph{Training process}
In our experiments we took the top $1M$ paths that occurred the most in the training set.
we use the Adam optimization algorithm~\cite{kingma2014adam}, an adaptive gradient descent method commonly used in deep learning. 
We use dropout \cite{srivastava2014dropout} of $0.25$ on the context vectors. The values of all the parameters are initialized using the initialization heuristic of \citet{glorot2010understanding}. When training on a single Tesla K80 GPU, we achieve a training throughput of more than 1000 methods per second. Therefore, a single training epoch takes about $3$ hours, and it takes about 1.5 days to completely train a model. Training on newer GPUs doubles and quadruples the speed.
Although the attention mechanism has the ability to aggregate an arbitrary number of inputs, we randomly sampled up to $k=200$ path-contexts from each training example. The value $k=200$ seemed to be enough to ``cover'' each method, since increasing to $k=300$ did not seem to improve the results.

\paragraph{Data sets}
We are interested in evaluating the ability of the approach to generalize across projects. We used a data set of $10,072$ Java GitHub repositories, originally introduced by \citet{pigeon-pldi}. Following recent work which found a large amount of code duplication in GitHub \cite{dejavu2017}, \citet{pigeon-pldi} used the top-ranked and most popular projects, in which duplication was observed to be less of a problem (\citet{dejavu2017} measured duplication across all the code in GitHub), and they filtered out migrated projects and forks of the same project. While it is possible that some duplications are left between the training and test set, in this case the compared baselines could have benefited from them as well. In this dataset, the files from all the projects are shuffled and split to $14,162,842$ training, $415,046$ validation and $413,915$ of test methods.

We trained our model on the training set, tuned hyperparameters on the validation set for maximizing F1 score.
The number of training epochs is tuned on the validation set using early stopping. Finally, we report results on the unseen test set.
A summary of the amount of data used is shown in \Cref{datasets}.

\begin{table}[]
\centering
\footnotesize
\begin{tabular}{lrrr}
\toprule
            &  Number of methods & Number of files & Size (GB)  \\
\midrule
Training    & 14,162,842        & 1,712,819    & 66          \\ 
Validation  & 415,046          & 50,000        & 2.3       \\ 
Test        & 413,915          & 50,000        & 2.3         \\
Sampled Test & 7,454          & 1,000        & 0.04         \\
\bottomrule
\end{tabular}
\caption{Size of data used in the experimental evaluation.}
\label{datasets}
\end{table}

\paragraph{Evaluation metric}
Ideally, we would like to manually evaluate the results, but given that manual evaluation is very difficult to scale, we we adopted the measure used by previous works \cite{conv16, pigeon-pldi, allamanis2015}, which measured precision, recall, and F1 score over sub-tokens, case-insensitive. This is based on the idea that the quality of a method name prediction is mostly dependant on the sub-words that were used to compose it. For example, for a method called \scode{countLines}, a prediction of \scode{linesCount} is considered as an exact match, a prediction of \scode{count} has a full precision but low recall, and a prediction of \scode{countBlankLines} has a full recall but low precision. An unknown sub-token in the test label (``UNK'') is counted as a false negative, therefore automatically hurting recall.

While there are alternative metrics in the literature, such as accuracy and BLEU score, they are problematic because accuracy counts even mostly-correct predictions as completely incorrect, and BLEU score tends to favor short predictions, which are usually uninformative \cite{callison2006re}.
We provide a qualitative evaluation including a manual inspection of examples in \Cref{qualitative}.

\subsection{Quantitative Evaluation}\label{quantitative}

We compare our model to two other recently-proposed models that address similar tasks:

\paragraph{CNN+attention} --- proposed by \citet{conv16} for prediction of method names using CNNs and attention. This baseline was evaluated on a random sample of the test set due to its slow prediction rate (\Cref{final_results}). We note that the results reported here are lower than the original results reported in their paper, because we consider the task of learning \emph{a single model that is able to predict names for a method from any possible project}. We do not make the restrictive assumption of having a per-project model, able to predict only names within that project. The results we report for CNN+attention are when evaluating their technique in this realistic setting. In contrast, the numbers reported in their original work are for the simplified setting of predicting names \emph{within the scope of a single project}.

\paragraph{LSTM+attention} --- proposed by \citet{codenn16}, originally for translation between StackOverflow questions in English and snippets of code that were posted as answers and vice-versa, using an encoder-decoder architecture based on LSTMs and attention. Originally, they demonstrated their approach for C\# and SQL. We used a Java lexer instead of the original C\#, and pedantically modified it to be equivalent. 
We re-trained their model with the target language being the methods' names, split into sub-tokens. Note that this model was designed for a slightly different task than ours, of translation between source code snippets and natural language descriptions, and not specifically for prediction of method names.

\paragraph{Paths+CRFs} --- proposed by \citet{pigeon-pldi}, using a similar syntactic path representation as this work, with CRFs as the learning algorithm. We evaluate our model on the their introduced dataset, and achieve a significant improvement in results, training time and prediction time.

Each baseline was trained on the same training data as our model. We used their default hyperparameters, except for the embedding and LSTM size of the LSTM+attention model, which were reduced from $400$ to $100$, to allow it to scale to our enormous training set while complying with the GPU's memory constraints. The alternative was to reduce the amount of training data, which achieved worse results.

\begin{table*}[]
\centering
\footnotesize
\begin{tabular}{llllllll}
\toprule
                                   & \multicolumn{3}{l}{Sampled Test Set (7454 methods)}  & \multicolumn{3}{l}{Full Test Set (413915 methods)} & prediction rate  \\
Model                              & Precision     & Recall        & F1            & Precision     & Recall        & F1           & (examples / sec) \\ \midrule
CNN+Attention \cite{conv16}     & 47.3          & 29.4          & 33.9          & -              & -              & -              & 0.1\\
LSTM+Attention \cite{codenn16}  & 27.5          & 21.5          & 24.1          & 33.7          & 22.0          &  26.6            & 5 \\
Paths+CRFs \cite{pigeon-pldi} & - & - & - & 53.6 & 46.6 & 49.9 & 10 \\
\textbf{PathAttention (this work)} & \textbf{63.3} & \textbf{56.2} & \textbf{59.5} & \textbf{63.1} & \textbf{54.4} & \textbf{58.4} & \textbf{1000}\\
\bottomrule
\end{tabular}
\caption{Evaluation comparison between our model and previous works.}
\label{final_results}
\end{table*}
\begin{figure*}
\begin{tikzpicture}[scale=1]
	\begin{axis}[
		xlabel={Training time (hours)},
		ylabel={F1 score},
        legend style={at={(0.97,0.15)},anchor=east,font=\tiny},
        xmin=0, xmax=74,
        ymin=0.0, ymax=60,
        xtick={0,6,...,72},
        ytick={0.0,5.0,...,60.0},
        grid = major,
        grid style={line width=0.1pt, draw=gray!10},
        major grid style={line width=.2pt,draw=gray!50}
    ]
	
    \addplot[color=blue, mark=*, line width=1pt] coordinates {
		(3,51.9)
		(6,55.8)
		(12,57.7)
		(18,57.8)
		(24,58.0)
        (30, 58.4)
        (72, 58.4)
	};
    \addlegendentry{\footnotesize PathAttention (this work)}

    \addplot[color=brown, mark=diamond*, mark size=3pt] coordinates {
        (12, 40.6)
        (24, 45.8)
        (36, 47.8)
		(48, 49.1)
        (60, 49.9)
		(72, 49.9)
	};
    \addlegendentry{\footnotesize Paths+CRFs}

    \addplot[color=red, mark=square*] coordinates {
		(12, 26.3)
		(24, 30.1)
        (48, 32.6)
        (72, 33.9)
	};
    \addlegendentry{\footnotesize CNN+Attention}
    \addplot[color=green, mark=triangle*] coordinates {
		(12, 17.7)
		(24, 19.5)
        (36, 21.5)
        (48, 23.7)
        (60, 24.6)
        (72, 25.2)
	};
    \addlegendentry{\footnotesize LSTM+Attention}

	\end{axis}

\end{tikzpicture}
    \caption{Our model achieves significantly higher results than the baselines and in shorter time. }
    \figlabel{main_results}
\end{figure*}
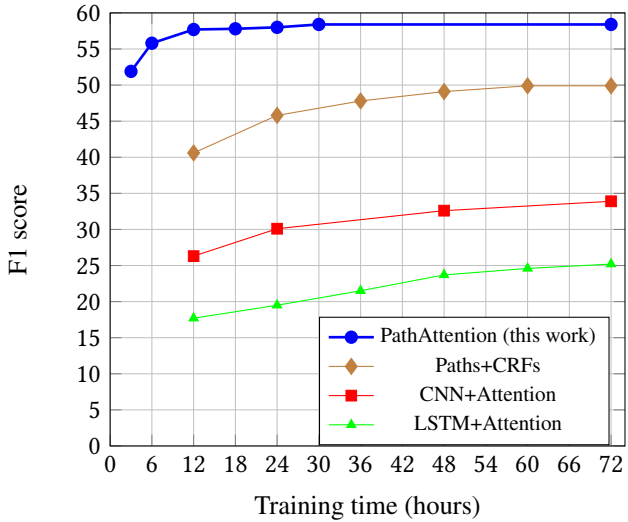 

\paragraph{Performance}
\Cref{final_results} shows the precision, recall, and F1 score of each model. The model of \citet{pigeon-pldi} seems to perform better than that of \citet{conv16} and \citet{codenn16}, while our model achieves significantly better precision and recall than all of them.

\paragraph{Short and long methods}
The reported results are based on evaluation on \emph{all} the test data. Additionally evaluating the performance of our model with respect to the length of a test method, we observe similar results across method lengths, with natural descent as length increases. For example, the F1 score of one-line methods is around $65$; for two-to-ten lines $59$; and for eleven-lines and further $52$, while the average method length is $7$ lines. We used all the methods in the dataset, regardless of their size. This shows the robustness of our model to the length of the methods. Short methods have shorter names and their logic is usually simpler, while long methods benefit from more context for prediction,
but their names are usually longer, more diverse and sparse, for example: \scode{generateTreeSetHashSetSpoofingSetInteger} which has $17$ lines of code.

\paragraph{Speed}
\figref{main_results} shows the test F1 score over training time for each of the evaluated models. In just $3$ hours, our model achieves results that are as $88\%$ as good as its final results, and in $6$ hours results that are as $95\%$ as good, while both being substantially higher than the best results of the baseline models. Our model achieves its best results after $30$ hours.

\Cref{final_results} shows the approximate prediction rate of the different models. The syntactic preprocessing time of our model is negligible but is included in the calculation.
As shown, due to their complexity and expensive beam search on prediction, the other models are several orders of magnitude slower than ours, limiting their applicability.

\paragraph{Data efficiency}
The results reported in \Cref{final_results} were obtained using our full and large training corpus, to demonstrate the ability of our approach to leverage enormous amounts of training data in a relatively short training time. However, in order to investigate the data efficiency of our model, we also performed experiments using smaller training corpora which are not reported in details here. With $20\%$ of the amounts of data, the F1 score of our model drops in only $50\%$. With $5\%$ of the data, the F1 score drops only to $30\%$ of our top results.
We do not focus on this series of experiments here, since our model can process more than a thousand of examples per second, so there is no significant practical point in deliberately limiting the size of the training corpus.

\subsection{Evaluation of Alternative Designs}\label{alternative}
We experiment with alternative model designs, in order to understand the contribution of each network component.

\paragraph{Attention}
As we refer to our approach as \emph{soft-attention}, we examine two other approaches which are the extreme alternatives to our approach:
\begin{enumerate}
\item \emph{No-attention} --- in which every path-context is given an \emph{equal} weight: the model uses the ordinary average of the path-contexts rather than learning a weighted average.
\item \emph{Hard-attention} --- in which instead of placing the attention ``softly'' over the path-contexts, all the attention is given to a single path-context, i.e., the network learns to select a \emph{single} most important path-context at a time.
\end{enumerate}

\begin{table}
\parbox{.45\linewidth}{
\footnotesize
\centering
\begin{tabular}{llll}
\toprule
Model Design            & Precision     & Recall        & F1            \\
\midrule
No-attention            & 54.4          & 45.3          & 49.4          \\
Hard-attention          & 42.1          & 35.4          & 38.5          \\
Train-soft, predict-hard & 52.7 & 45.9 & 49.1 \\
\textbf{Soft-attention} & \textbf{63.1} & \textbf{54.4} & \textbf{58.4} \\
\textbf{Element-wise soft-attention} & \textbf{63.7} & \textbf{55.4} & \textbf{59.3} \\
\bottomrule
\end{tabular}
\caption{Comparison of model designs.}
\label{soft-hard}
}
\end{table}

A new model was trained for each of these alternative designs. However, training hard-attention neural networks is difficult, because the gradient of the $argmax$ function is zero almost everywhere. Therefore, we experimented with an additional approach: \emph{train-soft, predict-hard}, in which training is performed using soft-attention (as in our ordinary model), and prediction is performed using hard-attention. \Cref{soft-hard} shows the results of all the compared alternative designs. As seen, hard-attention achieves the lowest results. This concludes that when predicting method names, or in general describing code snippets, it is more beneficial to use all the contexts with equal weights than focusing on the single most important one. \emph{Train-soft, predict-hard} improves over hard training, and gains similar results to no-attention. As soft-attention achieves higher scores than all of the alternatives, both on training and prediction, this experiment shows its contribution as a ``sweet-spot'' between no-attention and hard-attention.

\paragraph{Removing the fully-connected layer} To understand the contribution of each component of our model, we experiment with removing the fully connected layer (described in \Cref{NeuralNetwork}). In this experiment, soft-attention is applied directly on the \emph{context-vectors} instead of the \emph{combined context-vectors}. This experiment resulted in the same final F1 score as our regular model. Even though its training rate (training examples per second) was faster, it took more actual training time to achieve the same results. For example, instead of reaching results that are as $95\%$ as good as the final results in $6$ hours, it took $12$ hours, and a few more hours to achieve the final results than our standard model.

\paragraph{Element-wise soft-attention} We also experimented with \emph{element-wise soft-attention}. In this design, instead of using a single attention vector $\boldsymbol{a}\in\mathbb{R}^{d}$ to compute the attention for the whole combined context vector $\boldsymbol{\tilde{c}_i}$, there are $d$ attention vectors $\boldsymbol{a_1},...,\boldsymbol{a_{d}}\in\mathbb{R}^{d}$, and each of them is used to compute the attention for a different \emph{element}. Therefore, the attention weight for element $j$ of a combined context vector $\boldsymbol{\tilde{c}_{i}}$ is: $\text{attention weight }\alpha_{i_j} = \frac{\exp(\boldsymbol{\tilde{c}_{i}}^{T}\cdot \boldsymbol{a_j})}{\sum_{k=1}^{n}\exp(\boldsymbol{\tilde{c}_{k}}^{T}\cdot \boldsymbol{a_j})}$. 
This variation allows the model to compute a different attention score for each \emph{element} in the combined context vector, instead of computing the same attention score for the whole combined context vector.
This model achieved F1 score of $59.3$ (on the full test set) which is even higher than our standard soft-attention model, but since this model gives a different attention to different elements within the same context vector it is more difficult to interpret. Thus, this is an alternative model that gives slightly better results in the cost of losing its interpretability and slower training.

\subsection{Data Ablation Study}\label{data_study}

\begin{table}\parbox{.48\linewidth}{
\footnotesize
\centering
\begin{tabular}{lllll}
\toprule
\multicolumn{2}{l}{Path-context input} & Precision & Recall & F1 \\
\midrule
Full: &$\langle x_s,p,x_t\rangle$        & 63.1 & 54.4          & 58.4                              \\
Only-values: &$\langle x_s,\_\_,x_t\rangle$ & 44.9 & 37.1           & 40.6 \\
No-values: &$\langle\_\_,p,\_\_\rangle$  & 12.0 & 12.6         & 12.3 \\
Value-path: &$\langle x_s,p,\_\_\rangle$ & 31.5   & 30.1    & 30.7\\
One-value: &$\langle x_s,\_\_,\_\_\rangle$  & 10.6 & 10.4            & 10.7 \\
\bottomrule
\end{tabular}
\caption{Our model while hiding input components.}
\label{hide_components}
}
\end{table}
\paragraph{The contribution of each path-context element} To understand the contribution of each component of a path-context, we evaluate our best model on the same test set in the same settings, except that one or more input locations is ``hidden'' and replaced with a constant ``UNK'' symbol, such that the model cannot use this element for prediction. As the ``full'' representation is referred to as: $\langle x_s,p,x_t\rangle$, the following experiments were performed:
\begin{itemize}
\item ``only-values'' - using only the values of the terminals for prediction, without paths, and therefore representing each path-context as: $\langle x_s,\_\_,x_t\rangle$.
\item ``no-values'' - using only the path: $\langle\_\_,p,\_\_\rangle$, without identifiers and keywords.
\item ``value-path'' - allowing the model to use a path and one of its values: $\langle x_s,p,\_\_\rangle$.
\item ``one-value'' - using only one of the values: $\langle x_s,\_\_,\_\_\rangle$.
\end{itemize}
The results of these experiments are presented in \Cref{hide_components}. Interestingly, the ``full'' representation ($\langle x_s,p,x_t\rangle$) achieves better results than the sum of ``only-values'' and ``no-values'', without each of them alone ``covering'' for the other.
This shows the importance of using \emph{both} paths and keywords, and letting the attention mechanism learn how to combine them in every example. The lower results of ``only-values'' (compared to the full representation) show the importance of using syntactic paths. As shown in the table, dropping identifiers and keywords hurt the model more than dropping paths, but combining both of them achieves significantly better results.
``no-paths'' gets better results than ``no-values'', and ``single-identifiers'' gets the worst results.

The low results of ``no-words'' suggest that predicting names for methods with obfuscated names is a much more difficult task. In this scenario, it might be more beneficial to predict variable names as a first step using a model that was trained specifically for this task, and then predict a method name given the predicted variable names.

\subsection{Qualitative Evaluation}\label{qualitative}

\begin{figure}[t]
\begin{minipage}{7in}
\hspace{-0.3in}
\begin{tabular}{lll}
\begin{subfigure}[t]{0.355\textwidth}
\vskip 0pt
\vspace{-4.0mm}
\includegraphics[scale=0.09]{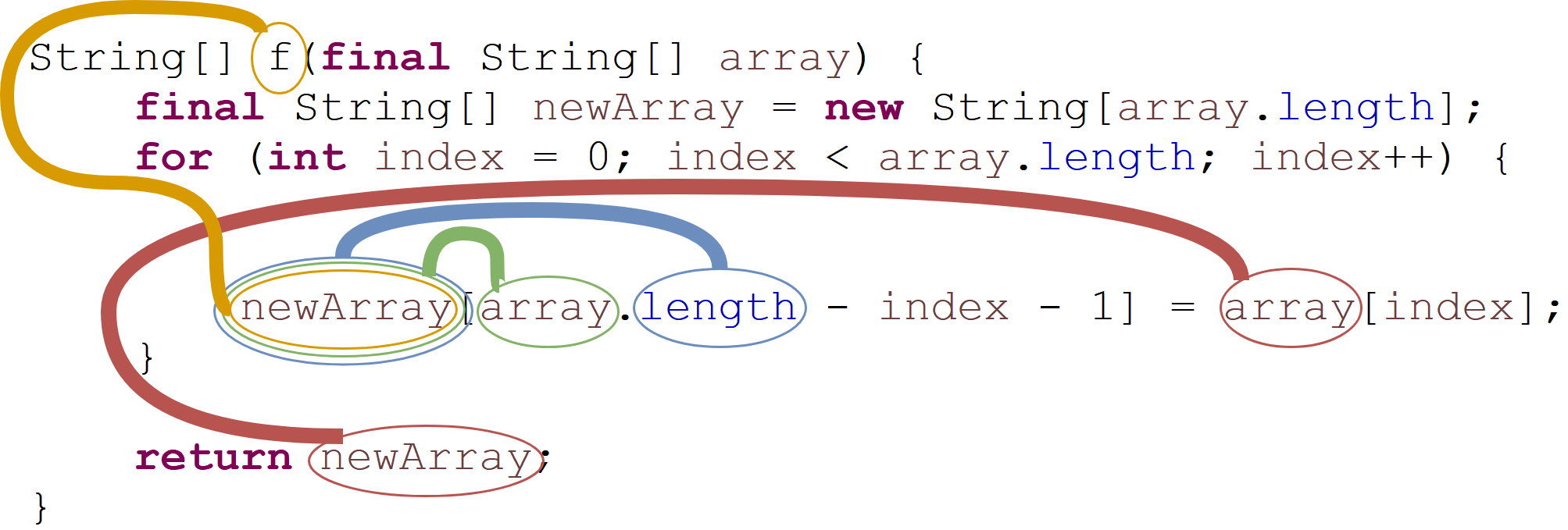}
\caption{}
\label{reverse}
\end{subfigure}
\rulesep
\begin{subfigure}[t]{0.185\textwidth}
\vskip 0pt
\vspace{-2.3mm}
\includegraphics[scale=0.045]{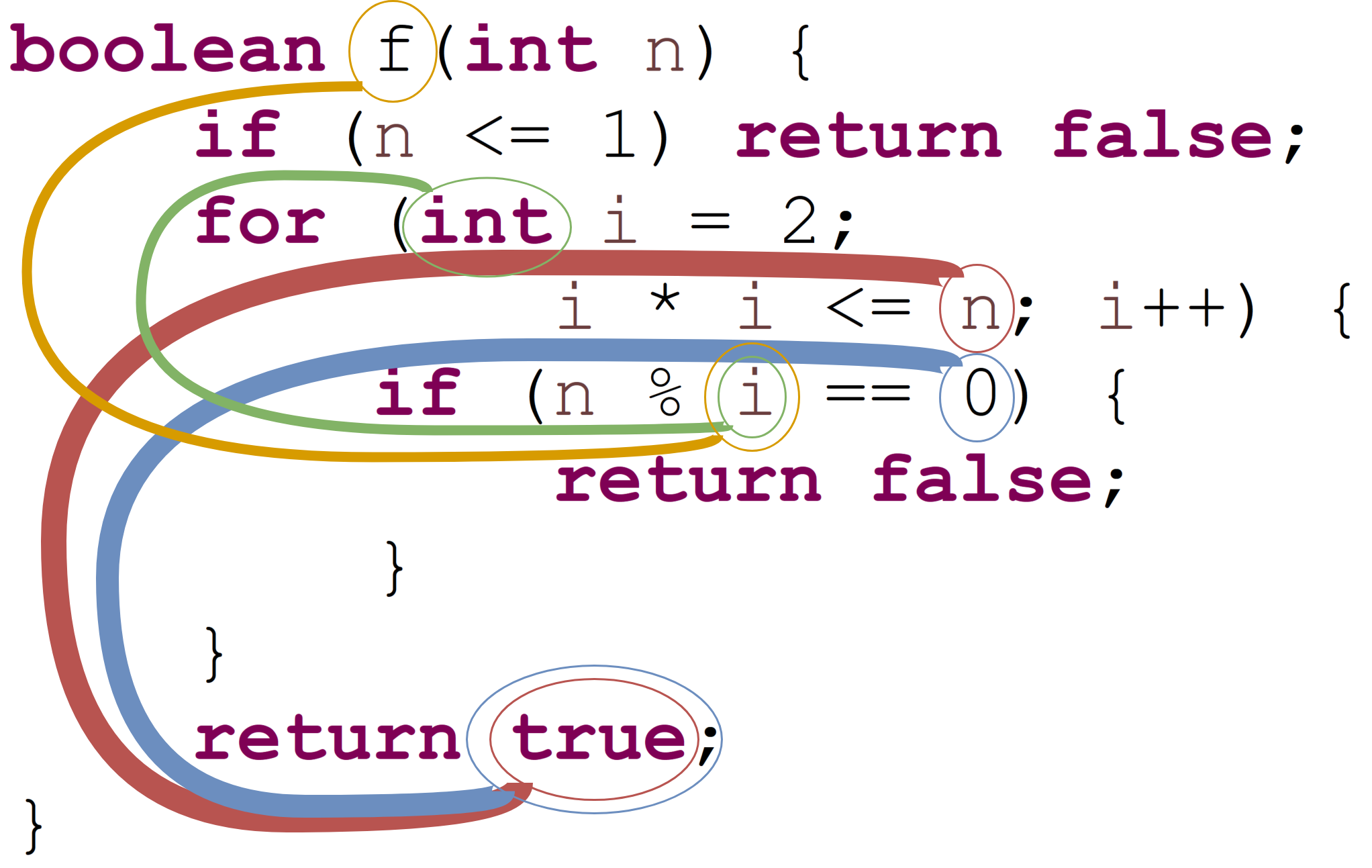}
\caption{}
\vspace{1mm}
\end{subfigure}
\rulesep
\begin{subfigure}[t]{0.4\textwidth}
\vskip 0pt
\vspace{-4.0mm}
\subcaptionbox{}{%
    \vspace{-19.5mm}
    \includegraphics[scale=0.09]{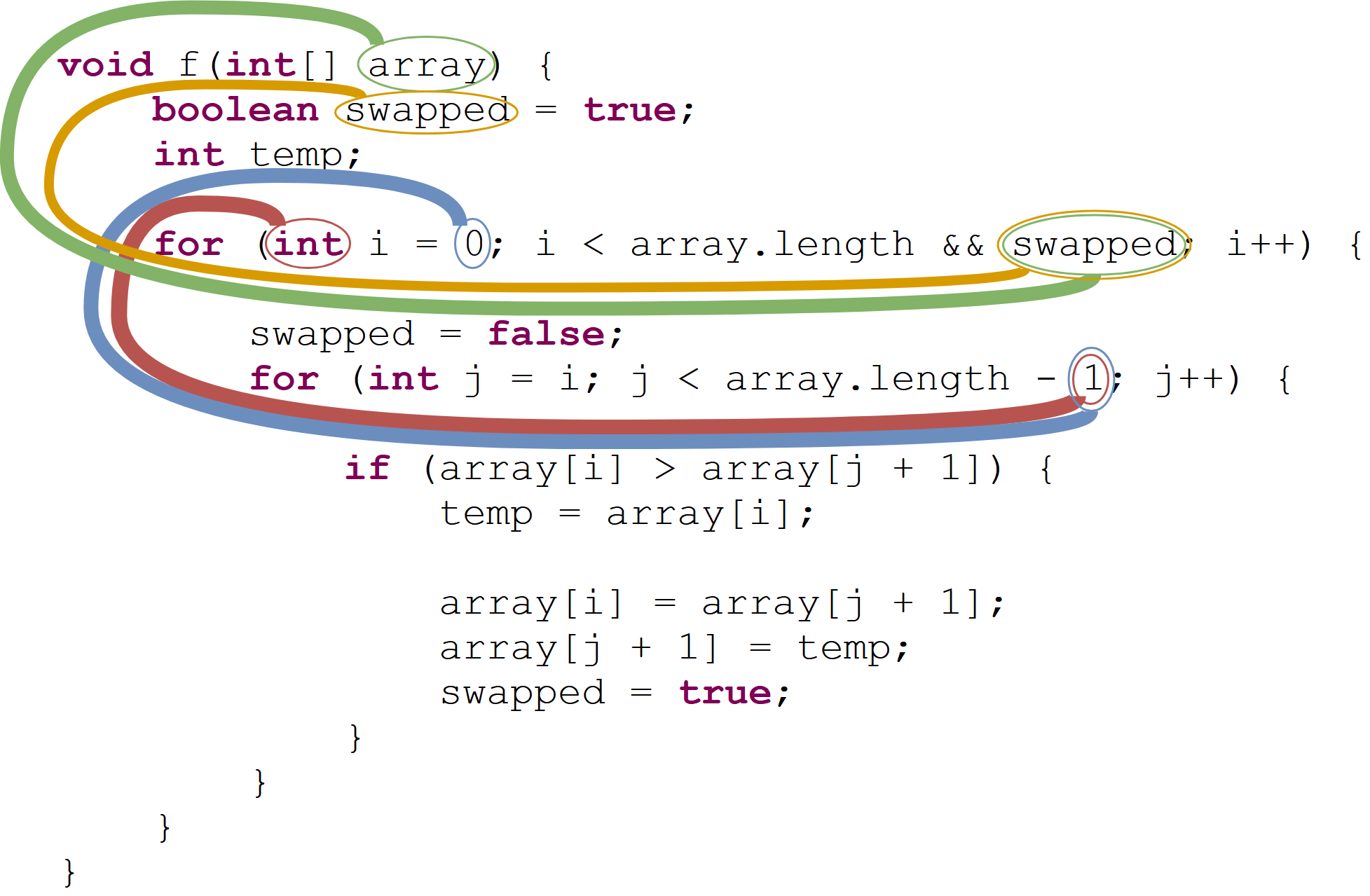}
\label{sort_example}
  }
\end{subfigure}
\\
\begin{subfigure}[t]{0.355\textwidth}
\vspace{-0.5mm}
\hspace{0.20in}
\scriptsize
\begin{tabular}{llr}
{\ul Predictions}  & &  \\
\textbf{reverseArray}   &\progressbar{0.7734}& $77.34\%$      \\
\textbf{reverse}        &\progressbar{0.1818}& $18.18\%$ \\
\textbf{subArray}       &\progressbar{0.0145}& $1.45\%$
\end{tabular}
\end{subfigure}
\rulesep
\begin{subfigure}[t]{0.185\textwidth}
\vspace{-0.5mm}
\scriptsize
\begin{tabular}{llr}
{\ul Predictions}  & &  \\
\textbf{isPrime}        &\progressbar{0.8844}& \\ 
\textbf{isNonSingular}  &\progressbar{0.0708}& \\ 
\textbf{factorial}      &\progressbar{0.0142}& \\ 
\end{tabular}
\end{subfigure}
\rulesep
\begin{subfigure}[t]{0.4\textwidth}
\vspace{-0.5mm}
\hspace{0.09in}
\scriptsize
\begin{tabular}{llr}
{\ul Predictions}  & &   \\
\textbf{sort}       &\progressbar{0.9980}& $99.80\%$      \\
\textbf{bubbleSort} &\progressbar{0.0013}& $0.13\%$      \\
\textbf{shorten}    &\progressbar{0.0002}& $0.02\%$      \\
\end{tabular}
\vspace{0.8mm}
\end{subfigure}

\end{tabular}
\end{minipage}
\caption{Example predictions from our model, with the top-4 attended paths for each code snippet. The width of each path is proportional to the attention it was given by the model.}
\label{eval-examples}
\end{figure} 

\subsubsection{Interpreting Attention}
\begin{figure*}
\begin{subfigure}[b]{0.65\textwidth}
\hspace{0.2in}
\includegraphics[scale=0.13]{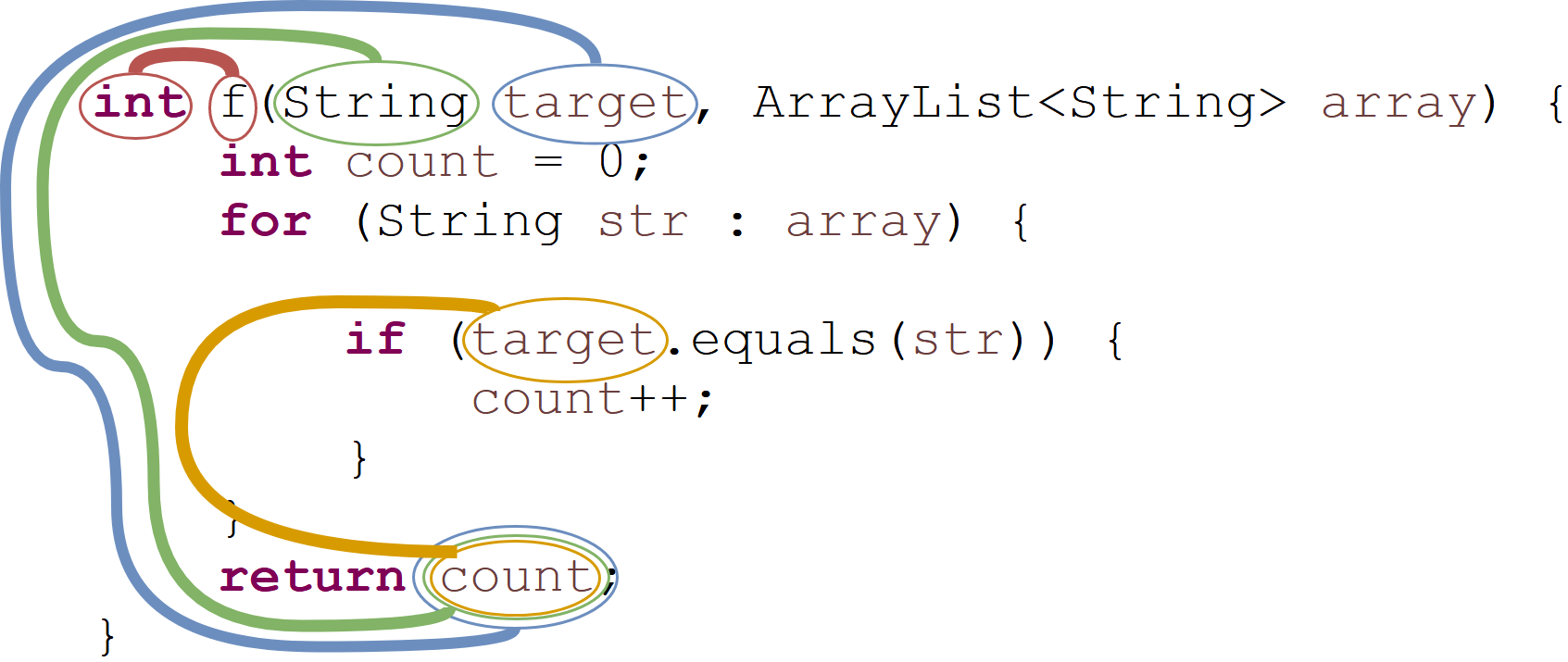}
\end{subfigure}
\begin{subfigure}[b]{0.3\textwidth}
\hspace{-2.5cm}
\begin{tabular}{lll}
{\ul Predictions}: & &\\
\textbf{count}              &\progressbar{0.4277}& $42.77\%$ \\
\textbf{countOccurrences}   &\progressbar{0.3374}& $33.74\%$ \\
\textbf{indexOf}            &\progressbar{0.0886}& $8.86\%$
\end{tabular}
\end{subfigure} \\


\begin{subfigure}[b]{1\textwidth}
\includegraphics[scale=0.11]{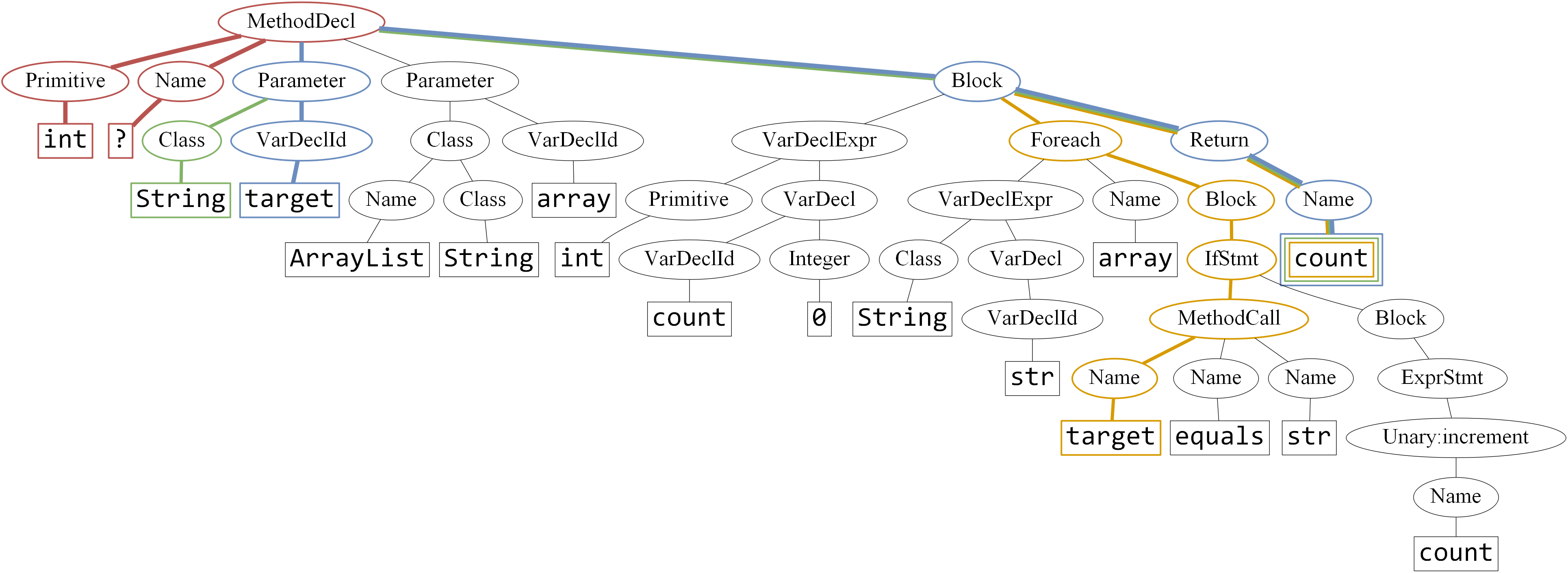}
\end{subfigure}

\caption{An example for a method name prediction, portrayed on the AST. The top-four path-contexts were given a similar attention, which is higher than the rest of the path-contexts.}
\label{appendix-count}
\end{figure*}

\begin{figure*}
\centering

\begin{subfigure}[b]{0.5\textwidth}
\hspace{0.2in}
\includegraphics[scale=0.13]{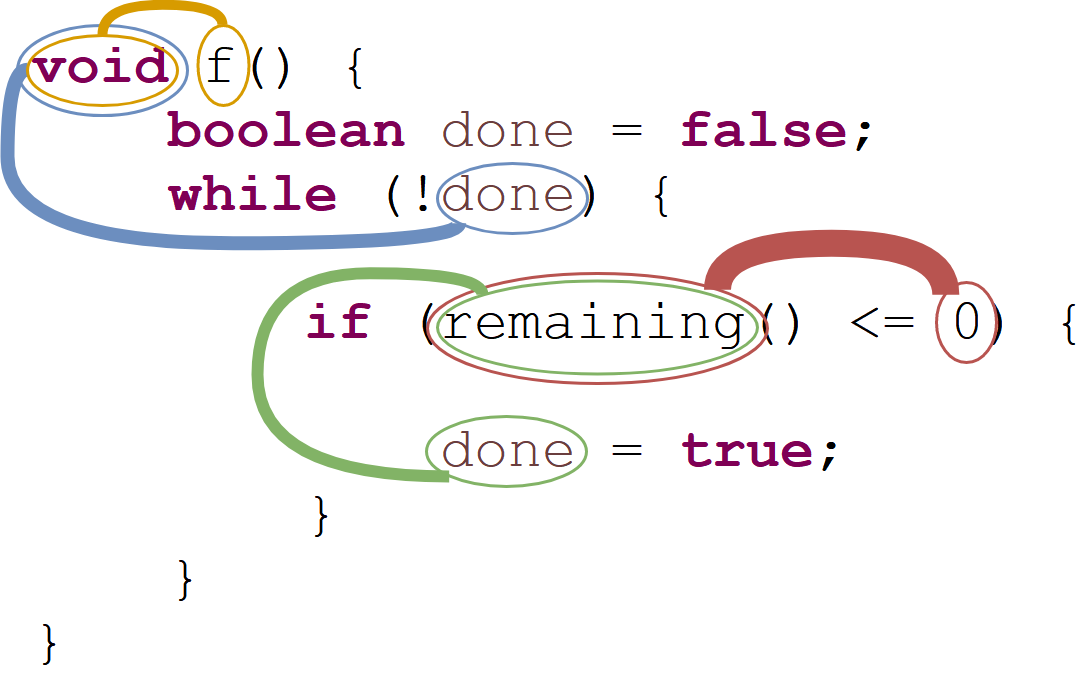}
\end{subfigure}
\begin{subfigure}[b]{0.4\textwidth}
\begin{tabular}{lll}
{\ul Predictions}: & &\\
\textbf{done}        &\progressbar{0.3427}& $34.27\%$ \\
\textbf{isDone}      &\progressbar{0.2979}& $29.79\%$ \\
\textbf{goToNext}    &\progressbar{0.1281}& $12.81\%$
\end{tabular}
\end{subfigure}


\begin{subfigure}[b]{1\textwidth}
\hspace{0.2in}
\includegraphics[scale=0.13]{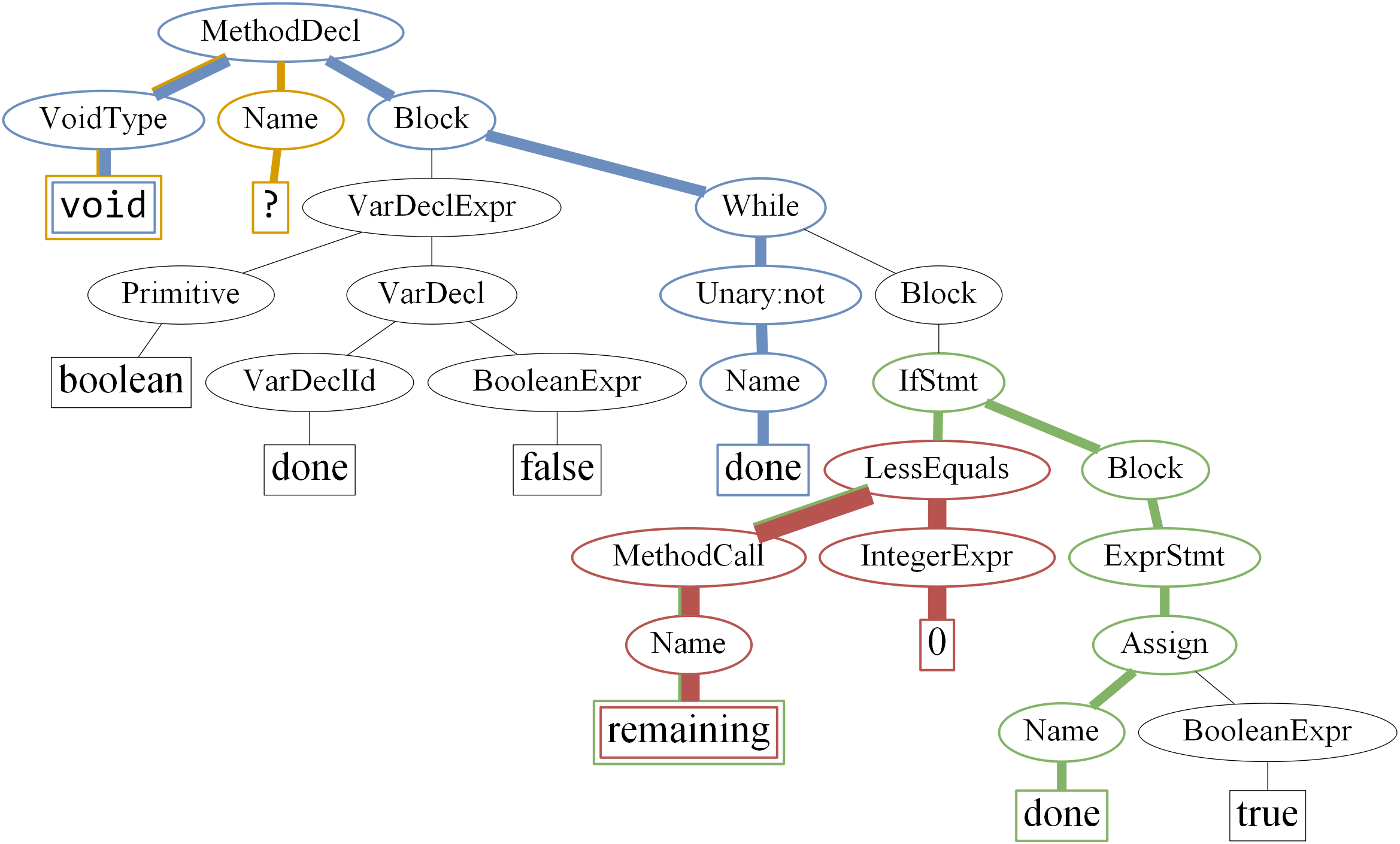}

\end{subfigure}
\caption{An example for a method name prediction, portrayed on the AST. The width of each path is proportional to the attention it was given.}
\label{appendix_done}
\end{figure*}

Despite the ``black-box'' reputation of neural networks, our model is partially interpretable thanks to the attention mechanism, which allows us to visualize the distribution of weights over the bag of path-contexts. \Cref{eval-examples} illustrates a few predictions, along with the path-contexts that were given the most attention in each method. The width of each of the visualized paths is proportional to the attention weight that it was allocated. We note that in these figures the path is represented only as a connecting line between tokens, while in fact it contains rich syntactic information which is not expressed properly in the figures. \Cref{appendix-count} and \Cref{appendix_done} portrays the paths on the AST.

The examples of \Cref{eval-examples} are particularly interesting since the top names are accurate and descriptive (\scode{reverseArray} and \scode{reverse}; \scode{isPrime}; \scode{sort} and \scode{bubbleSort}) but do not appear explicitly in the method bodies. The method bodies, and specifically the most attended path-contexts describe lower-level operations. Suggesting a descriptive name for each of these methods is difficult and might take time even for a trained human programmer.
The average method length in our dataset of real-world projects is $7$ lines, and the examples presented in this section are longer than this average length.



\Cref{appendix-count} and \Cref{appendix_done} show additional of our model's predictions, along with the path-contexts that were given the most attention in each example. The path-contexts are portrayed both on the code and on the AST.
An interactive demo of method name predictions and name vectors similarities can be found at: \url{http://code2vec.org}. When manually examining the predictions of custom inputs, it is important to note that a machine learning model learns to predict names for examples that are likely to be observed ``in the wild''. Thus, it can be misleaded by confusing adversarial examples that are unlikely to be found in real code.

\subsubsection{Semantic Properties of the Learned Embeddings}\label{semanticProperties}
Surprisingly, the learned method name vectors encode many semantic similarities and even analogies that can be represented as linear additions and subtractions.
When simply looking for the closest vector (in terms of cosine distance) to a given method name vector, the resulting neighbors usually contain semantically similar names; e.g. \scode{size} is most similar to \scode{getSize}, \scode{length}, \scode{getCount}, and \scode{getLength}. \Cref{similarities} shows additional examples of name similarities.

\begin{table}
\parbox{.47\linewidth}{
\footnotesize
\centering
\begin{tabular}{|l|l||l|}
\hline
A     & $+$B     & $\approx$C             \\ \hline
get  & value & getValue   \\
get  & instance & getInstance   \\
getRequest & addBody  & postRequest \\
setHeaders & setRequestBody & createHttpPost \\
remove & add & update \\
decode & fromBytes & deserialize \\
encode & toBytes & serialize \\
equals  & toLowerCase & equalsIgnoreCase   \\
\hline
\end{tabular}
\centering
\caption{Semantic combinations of method names.}
\label{additions}
}
\hfill
\parbox{.48\linewidth}{
\footnotesize
\centering
\begin{tabular}{|ll||ll|}
\hline
A : & B     & C :     & {\ul D}           \\ \hline
open :& connect     & close : &  {\ul disconnect } \\
key : &       keys          & value :  &  {\ul values }     \\
lower : &  toLowerCase & upper : &  {\ul toUpperCase} \\
down :  &  onMouseDown & up : &  {\ul onMouseUp} \\
warning :  & getWarningCount & error : & {\ul getErrorCount} \\
value :  & containsValue & key : & {\ul containsKey }\\
start :  &  activate  & end : & {\ul deactivate }\\
receive :  & download & send : &  {\ul upload }\\
\hline
\end{tabular}
\centering
\caption{Semantic analogies between method names.}
\label{analogies}
}
\end{table} 

When looking for a vector that is close to \emph{two} other vectors, we often find names that are semantic combinations of the two other names. Specifically, we can look for the vector $\boldsymbol{v}$ that maximizes the similarity to two vectors $\boldsymbol{a}$ and $\boldsymbol{b}$:
\begin{equation}
argmax_{\boldsymbol{v}\in V}\left(sim\left(\boldsymbol{a},\boldsymbol{v}\right)\,\circledast
\,sim\left(\boldsymbol{b},\boldsymbol{v}\right)\right)
\label{similarity1}
\end{equation}
where $\circledast$ is an arithmetic operator used to combine two similarities, and $V$ is a vocabulary of learned name vectors, $tags\_vocab$ in our case.
When measuring similarity using cosine distance, \Cref{similarity1} can be written as:
\begin{equation}
argmax_{\boldsymbol{v}\in V}\left(cos\left(\boldsymbol{a},\boldsymbol{v}\right)\circledast cos\left(\boldsymbol{b},\boldsymbol{v}\right)\right)
\label{similarity2}
\end{equation}

Neither $vec(\scode{equals})$ nor $vec(\scode{toLowerCase})$ are the closest vectors to $vec(\scode{equalsIgnoreCase})$ individually. However, assigning $\boldsymbol{a}=vec\left(\scode{equals}\right)$, $\boldsymbol{b}=vec\left(\scode{toLowerCase}\right)$ and using ``$+$'' as the operator $\circledast$, results with the vector of \scode{equalsIgnoreCase} as the vector that maximizes \Cref{similarity2} for $\boldsymbol{v}$. 

Previous work in NLP has suggested a variety of methods for combining similarities \cite{levy2014linguistic} for the task of natural language analogy recovery. Specifically, when using ``$+$'' as the operator $\circledast$, as done by \citet{mikolovDistributed2013}, and denoting $\boldsymbol{\hat{u}}$ as the unit vector of a vector $\boldsymbol{u}$, \Cref{similarity2} can be simplified to:
\begin{equation*}
argmax_{\boldsymbol{v}\in V}\left(
\boldsymbol{\hat{a}}+\boldsymbol{\hat{b}}
\right)\cdot \boldsymbol{\hat{v}}
\end{equation*}

Since cosine distance between two vectors equals to the dot product of their unit vectors. Particularly, this can be used as a simpler way to find the above combination of method name similarities:
\begin{equation*}
vec\left(\scode{equals}\right)+vec\left(\scode{toLowerCase}\right)\approx vec\left(\scode{equalsIgnoreCase}\right)
\end{equation*}
This implies that the model has learned that \scode{equalsIgnoreCase} is the most similar name to \scode{equals} \emph{and} \scode{toLowerCase} combined.
\Cref{additions} shows some of these examples.

Similarly to the way that syntactic and semantic word analogies were found using vector calculation in NLP by \citet{mikolovEfficient2013, mikolov2013linguistic}, the method name vectors that were learned by our model also express similar syntactic and semantic analogies. For example, $vec\left(\scode{download}\right) $-$ vec\left(\scode{receive}\right) $+$ vec\left(\scode{send}\right)$ results in a vector whose closest neighbor is the vector for \scode{upload}. This analogy can be read as: ``\scode{receive} is to \scode{send} as \scode{download} is to: \underline{\scode{upload}}''.
More examples are shown in \Cref{analogies}.

\section{Limitations of our model}
In this section we discuss some of the limitations of our model and raise potential future research directions.

\paragraph{Closed labels vocabulary} One of the major limiting factors is the closed label space we use as target - our model is able to predict only labels that were observed as-is at training time.
This works very well for the vast majority of targets (that repeat across multiple programs), but as the targets become very specific and diverse (e.g., \scode{findUserInfoByUserIdAndKey}) the model is unable to compose such names and usually catches only the main idea (for example: \scode{findUserInfo}). Overall, on a general dataset, our model outperforms the baselines by a large margin even though the baselines are technically able to produce complex names.

\paragraph{Sparsity and Data-hunger} There are three main sources of sparsity in our model:
\begin{itemize}
\item Terminal values are represented as whole symbols - e.g., each of \scode{newArray} and \scode{oldArray} is a unique symbol that has an embedding of its own, even though they share most of their characters (\scode{Array}).
\item AST paths are represented as monolithic symbols - two paths that share most of their AST nodes but differ in only a single node are represented as distinct paths which are assigned with distinct embeddings.
\item Target nodes are whole symbols, even if they are composed of more common smaller symbols.
\end{itemize}

These sources of sparsity make the model consume a lot of trained parameters to keep an embedding for each observed value. The large number of trained parameters results in a large GPU memory consumption at training time, increases the size of the stored model (about $1.4$ GB), and requires a lot of training data. Further, this sparseness potentially hurts the results, because modelling source code with a finer granularity of atomic units may have allowed the model to represent more unseen contexts as compositions of smaller atomic units, and would have increased the repetitiousness of atomic units across examples. In the model described in this paper, paths, terminal values or target values that were not observed in training time - cannot be represented. To address these limitations we train the model on a huge dataset of $14$M examples, but the model might not perform as well using smaller datasets. Although requiring a lot of GPU memory, training our model on millions of examples fits in the memory of a Tesla K80 GPU which is relatively old.

An alternative approach for reducing the sparsity of AST paths is to use \emph{path abstractions} where only parts of the path are used in the context (e.g., abstracting away certain kinds of nodes, merging certain kinds of nodes, etc.). 

\paragraph{Dependency on variable names} Since we trained our model on top-starred open-source projects where variable naming is usually good, the model has learned to leverage variable names to predict the target label. When given uninformative, obfuscated or adversarial variable names, the prediction of the label is usually less accurate. We are considering several approaches to address this limitation in future research. One potential solution is to train the model on a mixed dataset of good and hidden variable names, hopefully reducing model dependency on variable names; another solution is to apply a model that was trained for variable de-obfuscation first (such as \cite{pigeon-pldi, jsnice2015}) and feed the predicted variable names into our model. 
\section{Related Work}\seclabel{relatedWork}

\paragraph{Bimodal modelling of code and natural language} Several works have investigated the properties of source code as bimodal: it is at the same time executable for machines, and readable for humans \cite{murali2018bayou, codenn16, allamanis2015, conv16, allamanis2015bimodal, zilberstein2016}. This property drives the hope to model natural language conditioned on code and vice-versa. \citet{codenn16} designed a token-based neural model using LSTMs and attention for translation between source code snippets and natural language descriptions.
As we show in \Cref{Evaluation}, when trained for predicting method names instead of description, our model outperformed their model by a large gap. \citet{allamanis2015bimodal, maddison2014} also addressed the problem of translating between code and natural language, by considering the syntax of the code rather than representing it as a tokens stream.

\citet{conv16} have suggested a CNN for summarization of code. The main difference from our work is that they used attention over a ``sliding window'' of tokens, while our model leverages the syntactic structure of code and propose a simpler architecture which scales to large corpora more easily.
Their approach was mainly useful for learning and predicting code from the same project, and had worse results when trained a model using several projects. As we show in \Cref{Evaluation}, when trained on a multi-project corpus, our model achieved significantly better results.

\paragraph{Representation of code in machine learning models} A previous work suggested a general representation of program elements using syntactic relations \cite{pigeon-pldi}. They showed a simple representation that is useful across different tasks and programming languages: Java, JavaScript, Python and C\#, and therefore can be used as a default representation for any machine learning models for code. Our representation is similar to theirs, but can represent \emph{whole snippets of code}. Further, the main novelty of our work is the understanding that \emph{soft-attention over multiple contexts} is needed for embedding programs into a continuous space, and the use of this embedding to \emph{predict properties of a whole code snippet}.

Traditional machine learning algorithms such as decision trees \cite{decisionTrees2016}, Conditional Random Fields \cite{jsnice2015}, Probabilistic Context-Free Grammars \cite{phog16, gvero15, maddison2014, allamanis2014idioms}, n-grams \cite{allamanis2013, allamanis2014-conventions, Hindle:2012:NS:2337223.2337322, nguyen2013, raychev14} have been used for programming languages in the past. In \cite{DPY:PLDI17,DPY:PLDI16,DY14}, simple models are trained on various forms of~\emph{tracelets} extracted statically from programs. In \cite{KEY:POPL16, KRY:ASPLOS18}, language models are trained over sequences of API-calls extracted statically from binary code. 

\paragraph{Attention in machine learning} Attention models have shown great success in many NLP tasks such as neural machine translation \cite{bahdanau14, luong15, vaswani2017attention}, reading comprehension \cite{levy2017zeroshot, seo2016bidirectional, hermann2015}, and also in vision \cite{mnih2014, ba2014}, image captioning \cite{xu2015}, and speech recognition \cite{chorowski2015attention, bahdanau2016end}.
The general idea is to simultaneously learn to concentrate on a small portion of the input data and to use this data for prediction. \citet{xu2015} proposed the terms ``soft'' and ``hard'' attention for the task of image captioning.

Syntax-based contexts have been used by \citet{decisionTrees2016, phog16}.
Other than targeting different tasks, our work differs in two major aspects. First, these works traverse the AST only to identify a \emph{context node}, and do not use the information contained in the path itself. In contrast, our model uses the path itself as an input to the model, and can therefore generalize using this information when a known path is observed, even when the nodes in its ends have never been seen by the model before. Second,
our work differs in a major aspect that these models attempt to find a \emph{single most informative context} for each prediction. This approach resembles \emph{hard-attention}, in which the hardness is an inherent part of their model. In contrast, we suggest to use \emph{soft-attention}, that uses multiple contexts for prediction, with different weights for each. In previous works which used non-neural techniques, soft-attention is not even expressible.


\paragraph{Distributed representations} The idea of distributed representations of words date back to \citet{deerwester1990indexing} and even \citet{salton1975vector}, and are commonly based on the distributional hypothesis of \citet{harris1954distributional} and \citet{firth1957synopsis}, which states that words in similar contexts have similar meaning. These traditional methods included frequency-based methods, and specifically pointwise mutual information (PMI) matrices.

Recently, distributed representations of words, sentences, and documents \cite{doc2vec} were shown to help learning algorithms to achieve a better performance in a variety of tasks \cite{bengio2003, collobert2008, SocherEtAl2011:RNN, turian2010, glorot2011domain, turney2006similarity}. \citet{mikolovEfficient2013, mikolovDistributed2013} has introduced word2vec, a toolkit enabling the training of embeddings.
An analysis by \citet{levy2014neural} showed that word2vec's skip-gram model with negative sampling is implicitly factorizing a word-context PMI matrix, linking the modern neural approaches with traditional statistical approaches.

In this work, we use distributed representations of code elements, paths, and method names that are trained as part of our network. Distributed representations make our model generalize better, require \emph{less} parameters than methods based on symbolic representations, and produce vectors with the property that vectors of semantically similar method names are similar in the embedded space. 
\section{Conclusion}\seclabel{Conclusion}
We presented a new attention-based neural network for representing arbitrary-sized snippets of code using a learned fixed-length continuous vector. The core idea is to use soft-attention mechanism over syntactic paths that are derived from the Abstract Syntax Tree of the snippet, and aggregate all of their vector representations into a single vector.

As an example of our approach, we demonstrated it by predicting method names using a model that was trained on more than $14,000,000$ methods. In contrast with previous techniques, our model generalizes well and is able to predict names in files across different projects. We conjecture that the ability to generalize stems from the relative simplicity and the distributed nature of our model. Thanks to the attention mechanism, the prediction results are interpretable and provide interesting observations.

We believe that the attention-based model which uses a structural representation of code can serve as a basis for a wide range of programming language processing tasks. To serve this purpose, all of our code and trained model are publicly available at \url{https://github.com/tech-srl/code2vec}. 
\begin{acks}

We would like to thank Guy Waldman for developing the code2vec website (\url{http://code2vec.org}).
We also thank Miltiadis Allamanis and Srinivasan Iyer for their guidance in the use of their models in the evaluation section, and
Yaniv David, Dimitar Dimitrov, Yoav Goldberg, Omer Katz, Nimrod Partush, Vivek Sarkar and Charles Sutton for their fruitful comments.

The research leading to these results has received funding from the European Union's Seventh Framework Programme (FP7) under grant agreement no. 615688-ERC- COG-PRIME. Cloud computing resources were provided by an AWS Cloud Credits for Research award.

\end{acks}


\bibliography{bib}



\end{document}